%% file: IEEEtran.tex
\documentclass[10pt,journal,compsoc]{IEEEtran}
\input{math_commands.tex}

\usepackage{hyperref}
\usepackage{adjustbox}
\usepackage{booktabs}
\usepackage{graphicx}
\usepackage{subfigure}
\usepackage{amssymb}
\usepackage{algorithm,algorithmicx}
\usepackage{algpseudocode}
\usepackage{url}
\usepackage{multirow}
\usepackage{amsmath}
\usepackage{amsfonts}
\usepackage{times}
\usepackage{bm}
\usepackage{color}
\usepackage{stmaryrd}
\usepackage{soul}
\usepackage{threeparttable}
\usepackage{ragged2e}
\usepackage{xspace}
\usepackage[normalem]{ulem}
\setstcolor{red}

\definecolor{mygreen}{rgb}{0.1875,0.5976,0.1875}
\definecolor{myred}{rgb}{0.8,0.1875,0.1875}
\def\red{\textcolor{myred}}
\def\green{\textcolor{mygreen}}
\def\etal{{\em et al.\/}\,}
\newcommand{\ie}{\emph{i.e.,}\xspace}
\newcommand{\eg}{\emph{e.g.,}\xspace}

\renewcommand{\raggedright}{\leftskip=0pt \rightskip=0pt plus 0cm}
\newcommand{\tabincell}[2]{\begin{tabular}{@{}#1@{}}#2\end{tabular}}

\ifCLASSOPTIONcompsoc
\usepackage{cite}
\else
  \usepackage{cite}  
\fi

\ifCLASSINFOpdf
\else
\fi
\begin{document}

\title{Natural Language Video Localization: A Revisit in Span-based Question Answering Framework}

\author{Hao Zhang, Aixin Sun, Wei Jing, Liangli Zhen, Joey Tianyi Zhou and Rick Siow Mong Goh
\IEEEcompsocitemizethanks{
\IEEEcompsocthanksitem This research is supported by the Agency for Science, Technology and Research (A*STAR) under its AME Programmatic Funding Scheme (Project \#A18A1b0045 and \#A18A2b0046).
\IEEEcompsocthanksitem H.~Zhang is with the Institute of High Performance Computing, A*STAR, Singapore, 138632 and the School of Computer Science and Engineering, Nanyang Technological University, Singapore, 639798.
\IEEEcompsocthanksitem A.~Sun is with the School of Computer Science and Engineering, Nanyang Technological University, Singapore, 639798.
\IEEEcompsocthanksitem W.~Jing is with the Institute of Infocomm Research, A*STAR, Singapore, 138632.
\IEEEcompsocthanksitem L.~Zhen, J.T.~Zhou and R.S.M.~Goh are with the Institute of High Performance Computing, A*STAR, Singapore, 138632. 
\IEEEcompsocthanksitem Corresponding author: J.T.~Zhou (Email: joey.tianyi.zhou@gmail.com).}
}

\markboth{Accepted by IEEE Transactions on Pattern Analysis and Machine Intelligence (TPAMI)}{Shell \MakeLowercase{\textit{et al.}}: Bare Demo of IEEEtran.cls for Computer Society Journals}

\IEEEtitleabstractindextext{
\begin{abstract} 
\raggedright{Natural Language Video Localization (NLVL) aims to locate a target moment from an untrimmed video that semantically corresponds to a text query. Existing approaches mainly solve the NLVL problem from the perspective of computer vision by formulating it as ranking, anchor, or regression tasks. These methods suffer from large performance degradation when localizing on long videos. In this work, we address the NLVL from a new perspective, \ie span-based question answering (QA), by treating the input video as a text passage. We propose a video span localizing network (VSLNet), on top of the standard span-based QA framework (named VSLBase), to address NLVL. VSLNet tackles the differences between NLVL and span-based QA through a simple yet effective query-guided highlighting (QGH) strategy. QGH guides VSLNet to search for the matching video span within a highlighted region. To address the performance degradation on long videos, we further extend VSLNet to VSLNet-L by applying a multi-scale split-and-concatenation strategy. VSLNet-L first splits the untrimmed video into short clip segments; then, it predicts which clip segment contains the target moment and suppresses the importance of other segments. Finally, the clip segments are concatenated, with different confidences, to locate the target moment accurately. Extensive experiments on three benchmark datasets show that the proposed VSLNet and VSLNet-L outperform the state-of-the-art methods; VSLNet-L addresses the issue of performance degradation on long videos. Our study suggests that the span-based QA framework is an effective strategy to solve the NLVL problem.}
\end{abstract}

}

\maketitle

\IEEEdisplaynontitleabstractindextext
\IEEEpeerreviewmaketitle

\IEEEraisesectionheading{\section{Introduction}\label{sec:intro}}

\begin{figure}[t]
    \centering
    \includegraphics[width=0.48\textwidth]{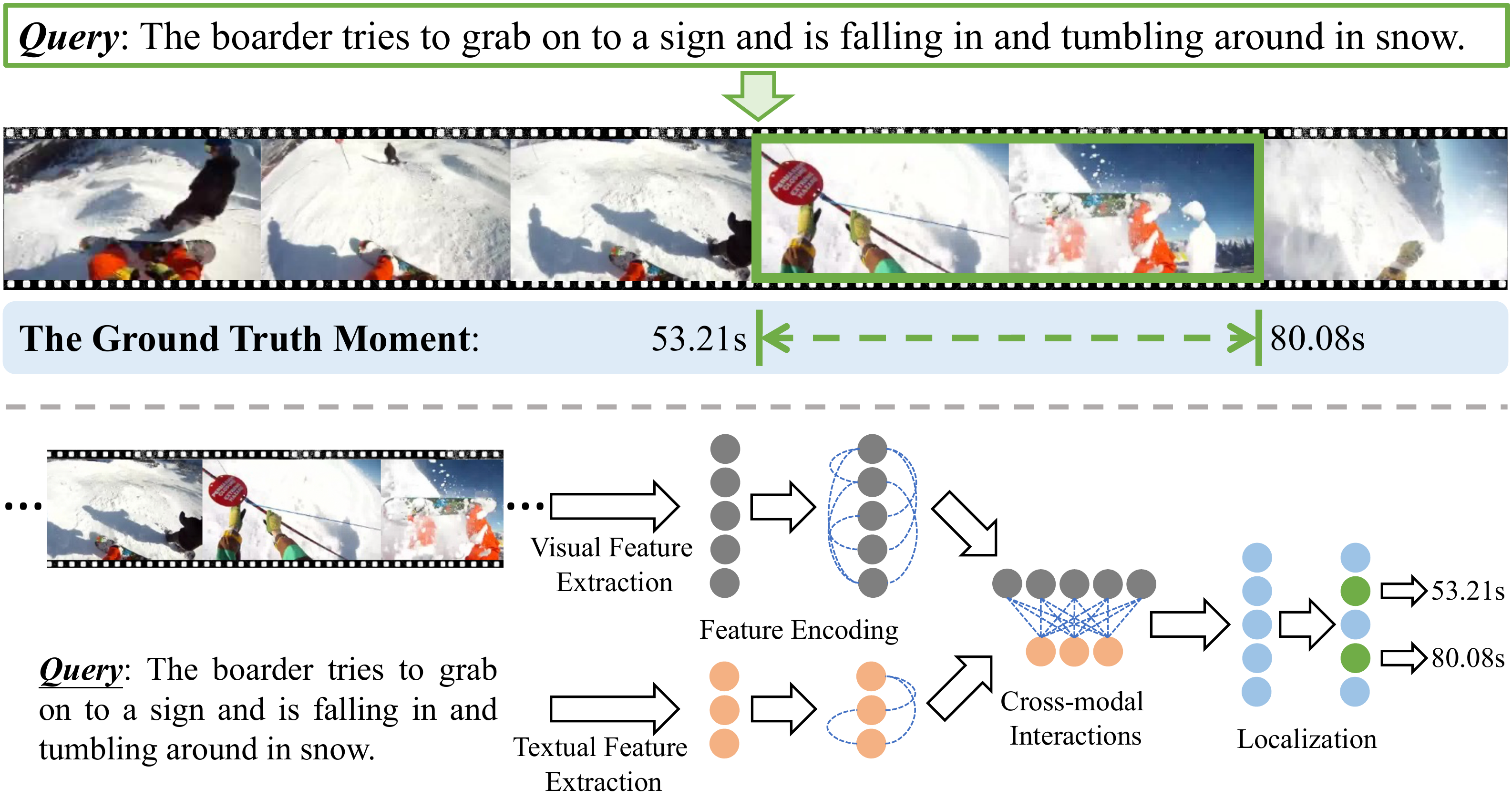}
    \caption{\small An illustration of localizing a temporal moment in an untrimmed video by a given language query, and the general procedures of natural language video localization.}
	\label{fig_example}
	\vspace{-0.3cm}
\end{figure}

\IEEEPARstart{N}{atural} language video localization (NLVL) is a prominent yet challenging problem in vision-language understanding. Given an untrimmed video, NLVL is to retrieve a temporal moment that semantically corresponds to a given language query. As illustrated in Figure~\ref{fig_example}, NLVL involves both computer vision and natural language processing techniques~\cite{Gao2018MotionAppearanceCN,yu2019activityqa,hu2019looking,le2019multimodal,zhou2020unified,lei2020tvqaplus}. Cross-modal reasoning is essential for NLVL to correctly locate the target moment in a video. Prior studies primarily treat NLVL as a ranking task, which apply multimodal matching architecture to find the best matching video segment for a query~\cite{hendricks2018localizing,Liu2018AMR,ge2019mac,Xu2019MultilevelLA,zhang2019man}. Some works~\cite{chen2018temporally,zhang2019man,zhu2019cross,Wang2020TemporallyGL} assign multi-scale temporal anchors to frames and select the anchor with the highest confidence as the result. Recently, several methods explore to model cross-interactions between video and query, and to regress temporal locations of target moment directly~\cite{Yuan2019ToFW,lu2019debug,chen2020rethinking}. There are also studies that formulate NLVL as a sequential decision-making problem and solve it with reinforcement learning~\cite{Wang2019LanguageDrivenTA,he2019Readwa,Wu2020TreeStructuredPB}.

Different from the aforementioned works, we address the NLVL task from a  new perspective, \ie span-based question answering (QA). Specifically, the essence of NLVL is to search for a video moment as the answer to a given language query from an untrimmed video. By treating the video as a text passage, and the target moment as the answer span, NLVL shares significant similarities with the span-based QA conceptually. With this intuition, NLVL could be revisited in the span-based QA framework.

However, the existing span-based QA methods could not be directly applied to solve the NLVL problem, due to the following two technical gaps. First, the data nature is different. To be specific, video is continuous, which results in the continuous causal relation inference between two consecutive video events; natural language, on the other hand, is discrete, and words in a sentence demonstrate syntactic structure. Therefore, changes between consecutive video frames are usually smooth. In contrast, two word tokens may carry different and even totally different meanings, \eg oxymoron or negation. As a result, events in a video are temporally correlated, and one can be linked with one another along the video sequence. The relationships between words or sentences are usually indirect and can be far apart. Second, small shifts in video frames are less imperceptible to humans than words in a text. Videlicet, small drifts between frames usually do not affect the understanding of video content. In contrast, the change of a few words or even one word could change the meaning of a sentence entirely.

By considering the above differences in data nature, we propose a \emph{video span localizing network} (\textbf{VSLNet}), where a \emph{query-guided highlighting (QGH)} strategy is introduced on the top of the traditional span-based QA framework~\cite{wei2018fast}. In QGH, we consider a region that covers the target moment by extending its starting and ending frames a bit further. In this way, the selected region is regarded  as foreground, while the rest is treated as background.

One challenge in NLVL is that the performance of many existing methods degrades significantly along with the increase of video length (see detailed discussion in Section~\ref{ssec:diff_video_len_exp}). Since the NLVL models generally perform well on short videos,  one straightforward solution to address this issue is to split a long video into multiple short clip segments. Then each clip segment is regarded as a short video. By treating a long video as a document, a clip segment as a paragraph, NLVL can be viewed as the multi-paragraph question answering (MPQA) task~\cite{clark2018simple}. The target moment in a long video can be considered as the answer span in a document for a given query.

However, how to properly split long video into clip segments is still challenging. Paragraphs in a document are semantically coherent units with boundaries defined by humans. Videos are continuous, and splitting the video into semantically coherent clip segments is difficult, even if it is feasible. In addition, the answer span in MPQA can be found in one of the paragraphs, but we cannot expect the target moment can be found within a single clip segment, regardless of how to split the videos. We propose a \emph{multi-scale split-and-concat strategy} to partition long video into clips of different lengths. Compared with fixed length splitting, the multi-scale splitting strategy increases the chance of locating a target moment in one segment. In this way, even if a target moment is truncated at one or several scales, segments in other scales may still be able to fully contain it. Thus, we can locate the moment in the clips that are more likely to contain it. This network is termed as \textbf{VSLNet-L}\footnote{``\textbf{L}'' represents the multi-scale split-and-concat strategy for \textbf{L}ong videos.} in the following context.

This paper is a substantial extension of our conference publication~\cite{zhang2020vslnet} with the following improvements. First, we investigate the localization performance degradation issue of existing NLVL models on long videos, and propose VSLNet-L to tackle this problem by introducing the concept of MPQA. Second, we carry out more experimental analyses involving the VSLNet-L and demonstrate its effectiveness on long videos. The contribution and novelty of this work are summarized as follows:
\begin{enumerate}
    \item We provide a new perspective to solve NLVL by formulating it as span-based QA, and analyze the natural differences between them in detail. To the best of our knowledge, this is one of the first works to adopt a span-based QA framework for NLVL.
    \item We propose VSLNet to explicitly address the differences between NLVL and span-based QA, by introducing a novel query-guided highlighting strategy.
    \item In addition to VSLNet, VSLNet-L is proposed by incorporating the concept of multi-paragraph QA to address the performance degradation on long videos.
\end{enumerate}

\begin{figure*}[t]
    \centering
    \includegraphics[width=1\textwidth]{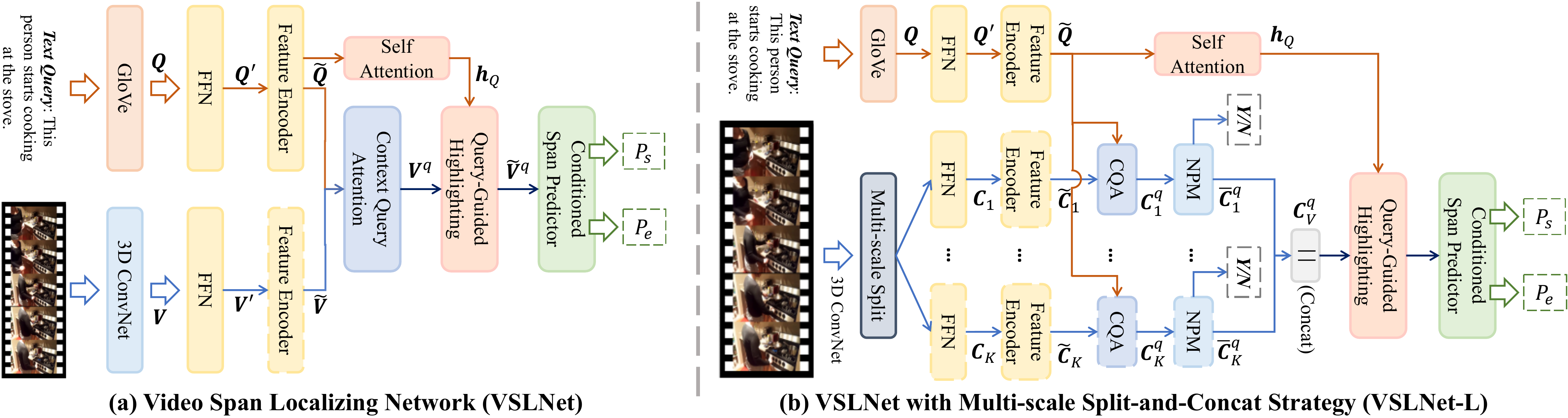}
    \vspace{-0.7cm}
    \caption{\small An overview of the proposed architectures for NLVL. The feature extractors, \ie GloVe and 3D ConvNet, are fixed during training. (a) depicts the structure of VSLNet. (b) shows the architecture of VSLNet-L. Note the standard span-based QA framework (VSLBase) is similar to VSLNet by removing the Self-Attention and Query-Guided Highlighting modules.}
	\label{fig_architecture}
\end{figure*}

\section{Related Work}\label{sec:related}

\subsection{Natural Language Video Localization}
As introduced by~\cite{Hendricks2017LocalizingMI,Gao2017TALLTA}, NLVL requires to model the cross-modality interactions between natural language texts and untrimmed videos, to retrieve video segments with language queries. Existing work on NLVL can be roughly divided into five categories: ranking, anchor, reinforcement learning, regression and span based methods.

Ranking-based methods~\cite{Hendricks2017LocalizingMI,hendricks2018localizing,Liu2018TemporalMN,Liu2018AMR,Liu2018CML,ge2019mac,Xu2019MultilevelLA,zhang2019exploiting} treat NLVL as a ranking task and rely on multimodal matching architectures to find the best matching video moment for a language query. For instance, Gao \etal~\cite{Gao2017TALLTA} proposed a cross-modal temporal regression localizer to jointly model the queries, as well as predicting alignment scores and action boundary regressions for pre-segmented candidate clips. Chen \etal~\cite{chen2019semantic} proposed a semantic activity proposal that integrates sentence queries' semantic information into the proposal generation process to get discriminative activity proposals. Although intuitive, these models are sensitive to negative samples. Specifically, they need densely sampled candidates to achieve good performance, leading to low efficiency and low flexibility.

Various approaches have been developed to overcome the above-mentioned drawbacks. Anchor-based methods~\cite{chen2018temporally,zhang2019man,zhu2019cross,Wang2020TemporallyGL,liu2020jointly,qu2020fine} sequentially process the video frames and assign each frame with multiscale temporal anchors; then the anchor with the highest confidence is selected as the result. For instance, Yuan \etal~\cite{yuan2019semantic} proposed a semantic conditioned dynamic modulation for better correlating sentence-related video contents over time and establishing a precise matching relationship between sentence and video. Zeng \etal~\cite{zeng2020dense} designed a dense regression network to regress the distances from each frame to the start/end frame of the video segment described by the query. Liu \etal~\cite{liu2020jointly} presented a cross- and self-modal graph attention network that recasts NLVL as a process of iterative messages passing over a joint graph to capture high-order interactions between two modalities.

Reinforcement learning-based methods formulate NLVL as a sequential decision-making problem~\cite{Wang2019LanguageDrivenTA,Wu2020TreeStructuredPB}. This process imitates humans' coarse-to-fine decision-making scheme to observe candidate moments conditioned on queries. For example, Wang \etal~\cite{Wang2019LanguageDrivenTA} proposed a recurrent neural network-based reinforcement learning model to selectively observe a sequence of frames and associate the given sentence with the video content in a matching-based manner. He \etal~\cite{he2019Readwa} presented a multi-task reinforcement learning framework to learn an agent that regulates the temporal grounding boundaries progressively based on its policy. Wu \etal~\cite{Wu2020TreeStructuredPB} presented a tree-structured policy-based progressive reinforcement learning framework to simulate humans' coarse-to-fine decision-making paradigm and to sequentially regulate the temporal boundary in an iterative refinement manner.

Regression-based methods~\cite{Yuan2019ToFW,chen2020rethinking} tackle NLVL by regressing the temporal time of target moments directly. Yuan \etal~\cite{Yuan2019ToFW} built a proposal-free model with BiLSTM and a three-step attention module to regress temporal locations of the target moment. Lu \etal~\cite{lu2019debug} proposed a dense bottom-up framework, which regresses the distances to start and end boundaries for each frame in the target moment. Then, it selects the one with the highest confidence as the result. Chen \etal~\cite{chen2020rethinking} further extended the bottom-up framework in~\cite{lu2019debug} with graph-based feature pyramid network to boost the performance. Mun \etal~\cite{mun2020local} proposed to extract a collection of semantic phrases in the query and reflect bi-modal interactions between the linguistic and visual features in multiple levels for moment boundaries regression.

Recently, a few attempts have been made to address NLVL that consider the concept of question answering~\cite{chen2019localizing,rodriguez2020proposal}. Specifically, Chen \etal~\cite{chen2019localizing} introduced a cross-gated attended recurrent network and a self interactor to exploit the fine-grained interactions between the query and video. A segment localizer is adopted to predict the starting and ending boundary of the moment. Ghosh \etal~\cite{ghosh2019excl} presented an extractive approach that predicts the start and end frames by leveraging cross-modal interactions between the text and video. However, they did not state and explain the similarity and differences between NLVL and the traditional span-based QA; and they did not adopt the standard span-based QA framework. In this work, we adopt a standard span-based QA framework to address NLVL; and propose VSLNet to explicitly address the issue caused by the differences between NLVL and the traditional span-based QA tasks.

In addition, several other methods have been applied to NLVL, Escorcia \etal~\cite{escorcia2019temporal} extends NLVL to a general case that the model is required to retrieve the target moments from a video corpus instead of from a single video. Shao \etal~\cite{Shao2018find} presented a unified framework to learn video-level matching and moment-level localization jointly. Mithun \etal~\cite{mithun2019weakly} explored to build a joint visual-semantic embedding based framework to learn latent alignment between video frames and sentence descriptions under weakly supervised setting. Lin \etal~\cite{Lin2020WeaklySupervisedVM} proposed a weakly-supervised moment retrieval framework requiring only coarse video-level annotations. Zhang \etal~\cite{zhang2020learning} modeled the temporal relations between video moments by a two-dimensional map and proposed a temporal adjacent network to encode discriminative features for matching video moments with referring expressions. Chen \etal~\cite{chen2020learning} presented a method for learning pairwise modality interactions to exploit complementary information better and improve performance on both temporal sentence localization and event captioning.

\subsection{Span-based Question Answering}
Span-based Question Answering (QA) has been widely studied in the past years. Wang \etal~\cite{wang2017machineCU} combined match-LSTM~\cite{wang2016learning} and Pointer-Net~\cite{vinyals2015pointer} to estimate boundaries of the answer span. BiDAF~\cite{Seo2017BidirectionalAF} introduced bi-directional attention to obtain query-aware context representation. Xiong \etal~\cite{Xiong2017DynamicCN} proposed a coattention network to capture the interactions between context and query. R-Net~\cite{wang2017gated} integrated mutual and self attentions into the RNN encoder for feature refinement. QANet~\cite{wei2018fast} leveraged on a similar attention mechanism in a stacked convolutional encoder to improve performance. FusionNet~\cite{huang2018fusionnet} presented a full-aware multi-level attention to capture complete query information. By treating input video as a text passage, the above frameworks are all applicable to NLVL in principle. However, these frameworks are not designed to consider the differences between video and text passage. Their modeling complexity arises because of the interactions between query and text passage, where the two inputs are in the homogeneous space. In our proposed method, VSLBase adopts a simple and standard span-based QA framework, making it easy to model the differences between video and text by adding additional modules. Our VSLNet addresses the issue caused by the differences with the QGH module.

Moreover, VSLNet-L is conceptually similar to the multi-paragraph question answering (MPQA) framework~\cite{clark2018simple,wang2018multipassage,wang2019multipassage,lin2019selecting}, where both methods explore to locate an answer span from multiple paragraphs/clips. Clark \etal~\cite{clark2018simple} adopted standard span-based QA models to entire documents scenario by introducing confidence modules to find an answer span from multiple paragraphs. Wang \etal~\cite{wang2018multipassage} proposed to solve MPQA by jointly training three modules that can predict the result based on answer boundary, answer content, and answer verification factors. Wang \etal~\cite{wang2019multipassage} presented a multi-passage BERT model to globally normalize answer scores across all passages of the same question. Lin \etal~\cite{lin2019selecting} proposed a learning to rank model with an attention module to select the best-matching paragraph for a question. Pang \etal~\cite{pang2019hasqa} presented a three-level hierarchical answer spans model to extract the answer from multi-paragraphs.

\section{Methodology}\label{sec:method}
In this section, we first describe how to address the NLVL task by adopting a span-based QA framework. Then, we present VSLBase (Sections~\ref{ssec:encoder} to~\ref{ssec:csp}) and VSLNet (Section~\ref{ssec:qgh}) in detail. Lastly, we detail VSLNet-L (Section~\ref{ssec:msc}) with multi-scale split-and-concat strategy. The model architectures are shown in Figure~\ref{fig_architecture}.

\subsection{Span-based QA for NLVL}\label{ssec:formulaion}
We denote the untrimmed video as $V=[f_1, f_2, \dots, f_T]$ and the language query as $Q=[q_1, q_2, \dots, q_m]$, where $T$ and $m$ are the number of frames and words, respectively. $\tau^{s}$ and $\tau^{e}$ represent the start and end time of the temporal moment \ie answer span. To address NLVL with the span-based QA framework, we transform the data into a set of SQuAD style triples $(Context, Question, Answer)$~\cite{rajpurkar2016squad}. For each video $V$, we extract its visual feature sequence $\bm{V}=[\bm{v}_1, \bm{v}_2, \dots, \bm{v}_n]$ by a pre-trained 3D ConvNet~\cite{Carreira2017QuoVA}, where $n$ is the number of extracted feature vectors. Here, $\bm{V}$ can be regarded as the sequence of word embeddings for a text passage with $n$ tokens, as in the traditional span-based QA framework. Similar to word embeddings, each feature vector $\bm{v}_i$ here is a video feature vector.

Since span-based QA aims to predict start and end boundaries of an answer span, the start/end time of a video sequence needs to be mapped to the corresponding boundaries in the visual feature sequence $\bm{V}$. Suppose the video duration is $\mathcal{T}$, the start (end) span index is calculated by $a^{s(e)}=\mathtt{Round}(\tau^{s(e)}/\mathcal{T}\times n)$, where $\mathtt{Round}(\cdot)$ denotes the rounding operator. During the inference, the predicted span boundary can be easily converted to the corresponding time via $\tau^{s(e)}=a^{s(e)}/n\times \mathcal{T}$. After transforming moment annotations in the NLVL dataset, we obtain a set of $(\bm{V},Q,\bm{A})$ triples. Visual feature sequence $\bm{V}=[\bm{v}_1, \bm{v}_2, \dots, \bm{v}_n]$ acts as the passage with $n$ tokens; $Q=[q_1, q_2,\dots, q_m]$ is the query with $m$ tokens, and the answer $\bm{A}=[\bm{v}_{a^{s}}, \bm{v}_{a^{s}+1}, \dots, \bm{v}_{a^{e}}]$ corresponds to a piece in the passage. Then, the NLVL task becomes to find the correct start and end boundaries of the answer span, $a^{s}$ and $a^{e}$.

\begin{figure}[t]
    \centering
    \includegraphics[trim={0cm 0cm 0cm 0cm},clip, width=0.17\textwidth, angle=-90]{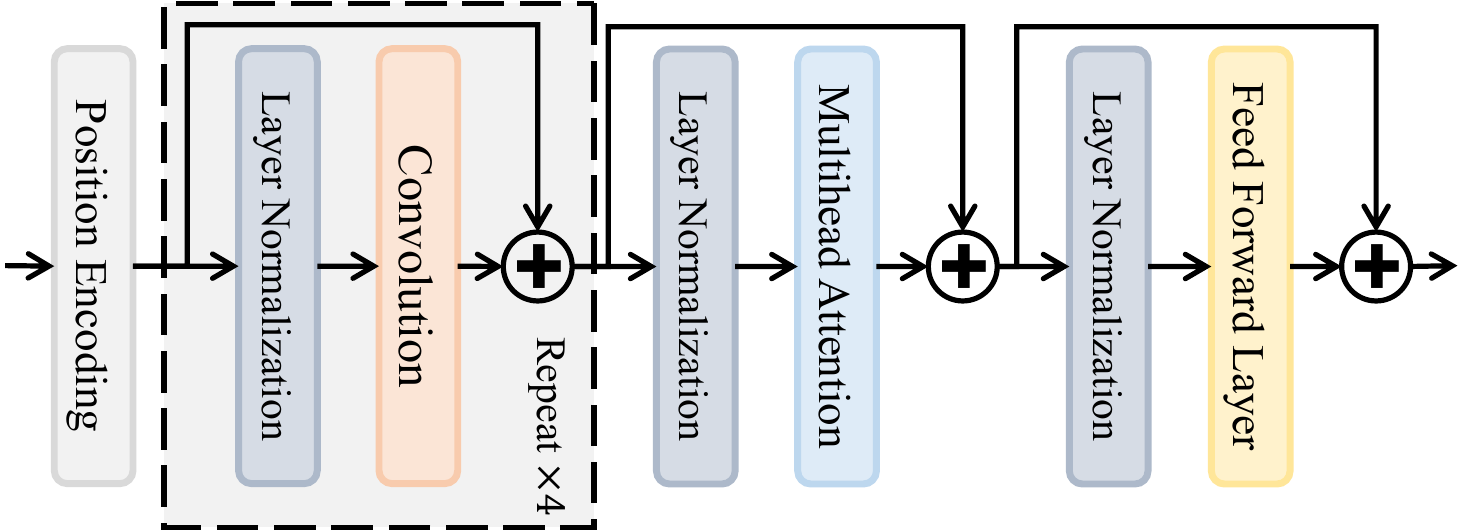}
	\caption{\small The structure of Feature Encoder.}
	\label{fig_encoder}
\end{figure}

\subsection{Feature Encoder}\label{ssec:encoder}
After obtaining visual features $\bm{V}=[\bm{v}_1,\bm{v}_2,\dots, \bm{v}_n]^{\top}\in\mathbb{R}^{n\times d_v}$, for a text query $Q$, we can compute its word embeddings $\bm{Q}=[\bm{q}_1,\bm{q}_2,\dots,\bm{q}_m]^{\top}\in\mathbb{R}^{m\times d_q}$ with existing text embedding approaches, \eg GloVe. We project the video and text feature vectors into the same dimension $d$, $\bm{V'}\in\mathbb{R}^{n\times d}$ and $\bm{Q'}\in\mathbb{R}^{m\times d}$, by two linear layers. Then we build the feature encoder with a simplified version of the embedding encoder layer in QANet~\cite{wei2018fast}.

Instead of applying a stack of multiple encoder blocks, we use only one encoder block, as shown in Figure~\ref{fig_encoder}. This encoder block consists of four convolution layers, followed by a multi-head attention layer~\cite{vaswani2017attention}. A feed-forward layer is used to produce the output. Layer normalization~\cite{Ba2016LayerN} and residual connection~\cite{he2016resnet} are applied to each layer. The encoded visual feature sequence and word embeddings are as follows:
\begin{equation}
\begin{aligned}
    \bm{\widetilde{V}} & =\mathtt{FeatureEncoder}(\bm{V'}) \\
    \bm{\widetilde{Q}} & =\mathtt{FeatureEncoder}(\bm{Q'})
\end{aligned}
\end{equation}
The parameters of feature encoder are shared by visual features and word embeddings.

\begin{figure}[t]
    \centering
    \includegraphics[width=0.48\textwidth]{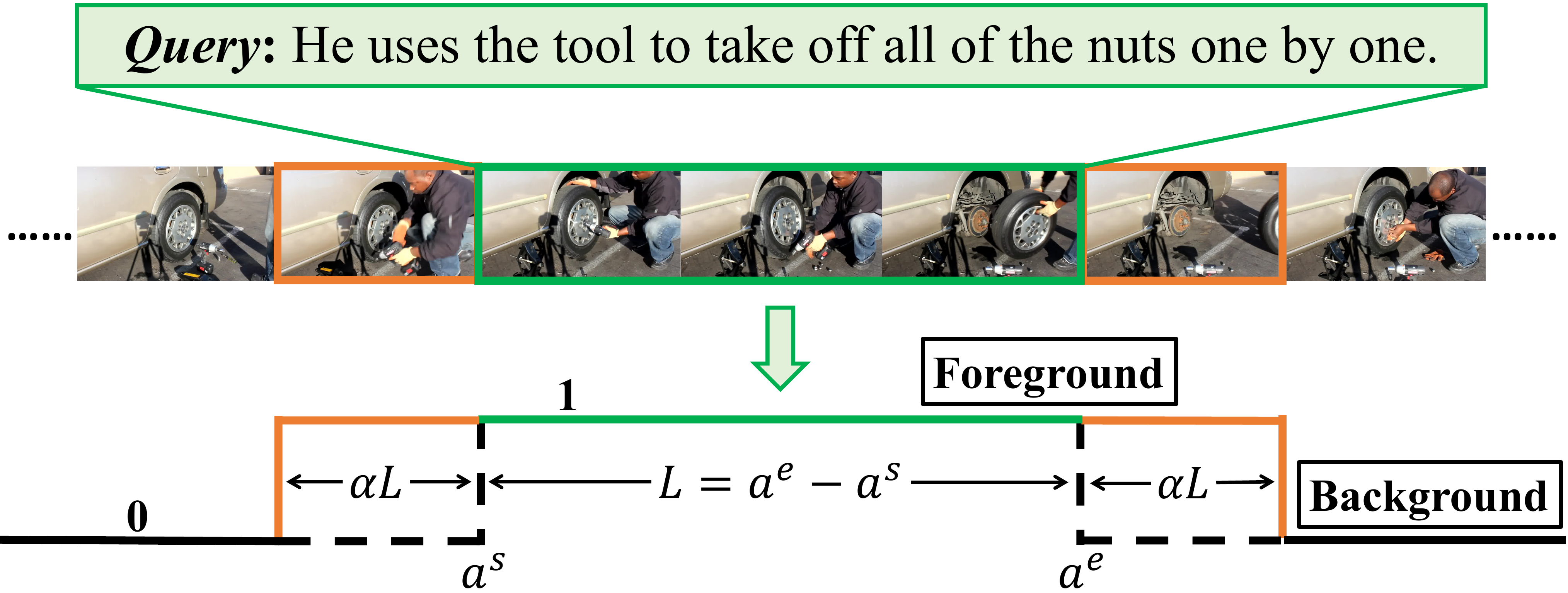}
    \vspace{-0.4cm}
    \caption{\small An illustration of foreground and background of visual features. $\alpha$ is the ratio of foreground extension.}
	\label{fig_highlight}
\end{figure}

\subsection{Context-Query Attention}\label{ssec:cqa}
After feature encoding, we utilize the context-query attention (CQA)~\cite{Seo2017BidirectionalAF,Xiong2017DynamicCN,wei2018fast} to capture the cross-modal interactions between visual and textural features. CQA first calculates the similarity scores, $\mathcal{S}\in\mathbb{R}^{n\times m}$, between each visual feature and query feature. Then context-to-query ($\mathcal{A}$) and query-to-context ($\mathcal{B}$) attention weights are computed as:
\begin{equation}
    \mathcal{A}=\mathcal{S}_{r}\cdot\bm{\widetilde{Q}}\in\mathbb{R}^{n\times d}, \mathcal{B}=\mathcal{S}_{r}\cdot\mathcal{S}_{c}^{T}\cdot\bm{\widetilde{V}}\in\mathbb{R}^{n\times d}
\end{equation}
where $\mathcal{S}_{r}$ and $\mathcal{S}_{c}$ are the row- and column-wise normalization of $\mathcal{S}$ by SoftMax, respectively. Finally, the output of context-query attention is written as:
\begin{equation}
    \bm{V}^{q}=\mathtt{FFN}\big([\bm{\widetilde{V}};\mathcal{A};\bm{\widetilde{V}}\odot\mathcal{A};\bm{\widetilde{V}}\odot\mathcal{B}]\big)
\end{equation}
where $\bm{V}^{q}\in\mathbb{R}^{n\times d}$; $\mathtt{FFN}$ is a single feed-forward layer; $\odot$ denotes an element-wise multiplication.

\begin{figure}[t]
    \centering
    \includegraphics[trim={0cm 0cm 0cm 0cm},clip, width=0.45\textwidth]{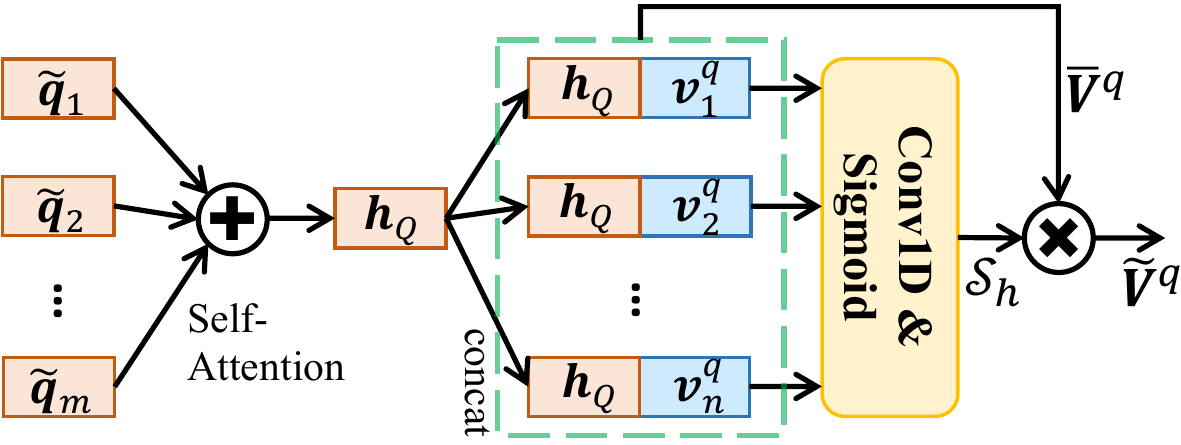}
    \vspace{-0.3cm}
	\caption{\small The structure of Query-Guided Highlighting (QGH).}
	\label{fig_qgh}
\end{figure}

\subsection{Conditioned Span Predictor}\label{ssec:csp}
We construct a conditioned span predictor using two unidirectional LSTMs and two feed-forward layers, inspired by Ghosh \etal~\cite{ghosh2019excl}. The main difference between our method and the method in~\cite{ghosh2019excl} is that we use unidirectional LSTM instead of bidirectional LSTM. We observe that unidirectional LSTM shows similar performance but with fewer parameters and higher efficiency. We believe that the unidirectional LSTM works as well as bidirectional LSTM in this setting for two possible reasons. First, video is a temporal sequence where  events happen earlier lead to later events, but not in the other way round. Second, the end boundary of a target moment always appears after the start boundary in the sequence. Therefore, we stack the two LSTMs so that the LSTM for end boundary is conditioned on the LSTM for start boundary, to maintain the sequence nature of video. The hidden states of the two LSTMs are then fed into the corresponding feed-forward layers to compute the start and end scores:
\begin{equation}
\begin{aligned}
    \bm{h}_{t}^{s} & =\mathtt{UniLSTM}_\textrm{start}(\bm{v}_{t}^{q}, \bm{h}_{t-1}^{s}) \\
    \bm{h}_{t}^{e} & =\mathtt{UniLSTM}_\textrm{end}(\bm{h}_{t}^{s}, \bm{h}_{t-1}^{e}) \\
    \bm{S}_{t}^{s} & = \bm{W}_{s}([\bm{h}_{t}^{s};\bm{v}_{t}^{q}]) + \bm{b}_{s} \\
    \bm{S}_{t}^{e} & = \bm{W}_{e}([\bm{h}_{t}^{e};\bm{v}_{t}^{q}]) + \bm{b}_{e}
\end{aligned}\label{eq_predictor}
\end{equation}
where $\bm{S}_{t}^{s/e}\in\mathbb{R}^{n}$ denote the scores of start/end boundaries at position $t$; $\bm{v}_{t}^{q}$ represents the $t$-th element in $\bm{V}^{q}$. $\bm{W}_{s/e}$ and $\bm{b}_{s/e}$ denote the weight matrix and bias of the start/end feed-forward layer, respectively.
Then, the probability distributions of start/end boundaries are computed by $P_{s/e} = \textrm{SoftMax}(\bm{S}^{s/e})\in\mathbb{R}^{n}$, and the training objective is defined as follows:
\begin{equation}
    \mathcal{L}_{\textrm{span}} = \frac{1}{2}\big[f_{\textrm{CE}}(P_{s}, Y_{s}) + f_{\textrm{CE}}(P_{e}, Y_{e})\big]
\end{equation}
where $f_{\textrm{CE}}$ is the cross-entropy loss function; $Y_{s}$ and $Y_{e}$ are the labels for the start ($a^{s}$) and end ($a^{e}$) boundaries, respectively. During the inference, the predicted answer span $(\hat{a}^{s},\hat{a}^{e})$ of a query is generated by maximizing the joint probability of start and end boundaries as:
\begin{equation}
\begin{aligned}
    \mathtt{span}(\hat{a}^{s},\hat{a}^{e}) & = \arg\max_{\hat{a}^{s},\hat{a}^{e}} P_{s}(\hat{a}^{s}) P_{e}(\hat{a}^{e}) \\
    \textrm{s.t. } & 0\leq\hat{a}^{s}\leq \hat{a}^{e}\leq n
\end{aligned}\label{eq:infer}
\end{equation}

The description of VSLBase architecture has been completed. VSLNet is built on top of VSLBase with QGH, to be detailed in Subsection~\ref{ssec:qgh}.

\begin{figure}[t]
    \centering
	\subfigure[\small An ideal case of splitting video into clip segments.]
	{
	    \label{fig_split_ideal}	
	    \includegraphics[trim={0cm 0cm 0cm 0cm},clip,width=0.45\textwidth]{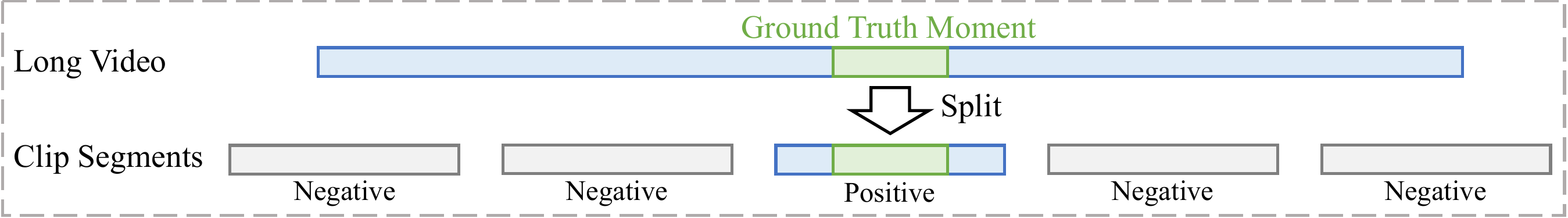}
	}
    \subfigure[\small A non-ideal case of splitting video into clip segments.]
	{
	    \label{fig_split_special}	
	    \includegraphics[trim={0cm 0cm 0cm 0cm},clip, width=0.45\textwidth]{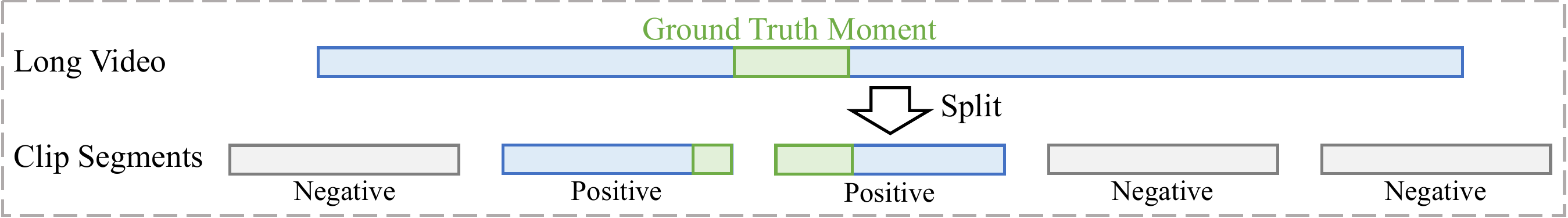}
	}
	\subfigure[\small An example of splitting video with two different scales.]
	{
	    \label{fig_split_multi_scale}	
	    \includegraphics[trim={0cm 0cm 0cm 0cm},clip, width=0.45\textwidth]{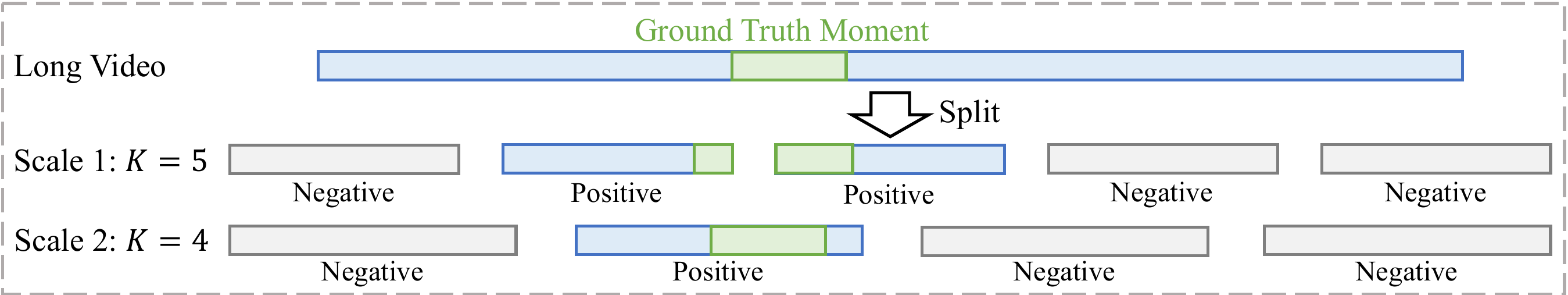}
	}
	\vspace{-0.2cm}
	\caption{\small Illustration of splitting video into clip segments.}
	\label{fig_split_demo}
\end{figure}

\subsection{Query-Guided Highlighting}\label{ssec:qgh}
The Query-Guided Highlighting (QGH) strategy is introduced in VSLNet, to address the major differences between text span-based QA and NLVL tasks, as shown in Figure~\ref{fig_architecture}(a). With the QGH strategy, we consider the target moment as the foreground, and the rest as background, illustrated in Figure~\ref{fig_highlight}. The target moment, which is aligned with the language query, starts from $a^{s}$ and ends at $a^{e}$ with length $L=a^{e}-a^{s}$. QGH extends the boundaries of the foreground to cover its antecedent and consequent video contents, where the extension ratio is controlled by a hyperparameter $\alpha$. As aforementioned in Introduction, the extended boundary could potentially cover additional contexts and also help the network to focus on subtle differences among video frames. 

By assigning $1$ to the foreground and $0$ to the background, we obtain a binary sequence, denoted as $Y_{\textrm{h}}$. QGH is a binary classification module to predict the confidence a visual feature vector belongs to the foreground or background.
The structure of QGH is shown in Figure~\ref{fig_qgh}. We first encode word feature representations $\bm{\widetilde{Q}}$ into sentence representation (denoted by $\bm{h}_{Q}$), with self-attention mechanism~\cite{Bahdanau2015NeuralMT}.
Then $\bm{h}_{Q}$ is concatenated with each feature element in $\bm{V}^{q}$ as $\bm{\bar{V}}^{q}=[\bm{\bar{v}}_{1}^{q},\dots,\bm{\bar{v}}_{n}^{q}]$, where $\bm{\bar{v}}_{i}^{q}=[\bm{v}_{i}^{q};\bm{h}_{Q}]$.
The highlighting score and highlighted feature representations are computed as:
\begin{equation}
\begin{aligned}
    \mathcal{S}_\textrm{h} & =\sigma\big(\mathtt{Conv1D}(\bm{\bar{V}}^{q})\big) \\
    \bm{\widetilde{V}}^{q} &=\mathcal{S}_\textrm{h}\cdot\bm{\bar{V}}^{q}
\end{aligned}\label{eq:qgh}
\end{equation}
where $\mathcal{S}_{\textrm{h}}\in\mathbb{R}^{n}$; $\sigma$ denotes the Sigmoid activation; $\cdot$ represents multiplication operator, which multiplies each feature of $\bar{V}^q$ and the corresponding score of $\mathcal{S}_\textrm{h}$. Accordingly, feature $\bm{V}^{q}$ in Equation~\ref{eq_predictor} is replaced by $\bm{\widetilde{V}}^{q}$ in VSLNet to compute  $\mathcal{L}_{\textrm{span}}$. The loss function of query-guided highlighting is formulated as:
\begin{equation}
    \mathcal{L}_{\textrm{QGH}}=f_{\textrm{CE}}(\mathcal{S}_{\textrm{h}}, Y_{\textrm{h}})
\end{equation}
VSLNet is trained in an end-to-end manner by minimizing the following overall loss:
\begin{equation}
    \mathcal{L}=\mathcal{L}_{\textrm{span}} + \mathcal{L}_{\textrm{QGH}}
\end{equation}

\subsection{Multi-scale Split-and-Concat}\label{ssec:msc}
To address the localization performance degradation issues on long videos, we introduce VSLNet-L with a multi-scale split-and-concatenation strategy. The architecture of VSLNet-L is shown in Figure~\ref{fig_architecture}(b). 

\begin{figure}[t]
    \centering
    \includegraphics[trim={0cm 0cm 0cm 0cm},clip, width=0.4\textwidth]{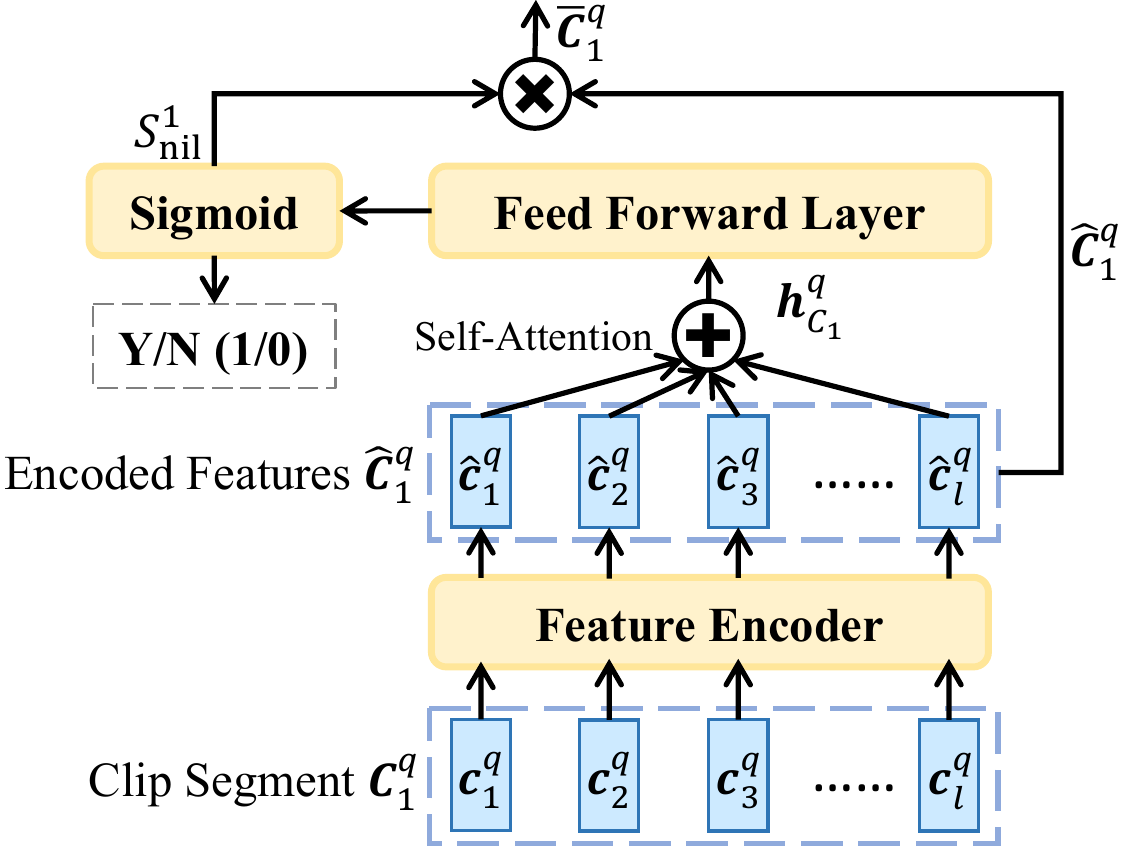}
    \vspace{-0.2cm}
	\caption{\small The structure of Nil Prediction Module (NPM).}
	\label{fig_npm}
\end{figure}

\begin{table*}[t]
    \small
    \caption{\small Statistics of the Natural Language Video Localization (NLVL) datasets.}
    \vspace{-0.3cm}
    \setlength{\tabcolsep}{5.0pt}
	\centering
	\begin{threeparttable}
	\begin{tabular}{ l c c c r r r r r }
		\toprule
		Dataset & Domain & \# Videos (train/val/test) & \# Annotations & $N_{\textrm{vocab}}$ & $\bar{L}_{video}$ & $\bar{L}_{query}$ & $\bar{L}_{moment}$ & $\Delta_{moment}$  \\
		\midrule
        Charades-STA & Indoors & $5,338/-/1,334$ & $12,408/-/3,720$ & $1,303$ & $30.59s$ & $7.22$ & $8.22s$ & $3.59s$ \\
        \midrule
        ANetCap & Open & $10,009/-/4,917$ & $37,421/-/17,505$ & $12,460$ & $117.61s$ & $14.78$ & $36.18s$ & $40.18s$ \\
        \midrule
        TACoS$_{\text{org}}$ & Cooking & $75/27/25$ & $10,146/4,589/4,083$ & $2,033$ & $287.14s$ & $10.05$ & $5.45s$ & $7.56s$ \\
        \midrule
        TACoS$_{\text{tan}}$ & Cooking & $75/27/25$ & $9,790/4,436/4,001$ & $1,983$ & $287.14s$ & $9.42$ & $25.26s$ & $50.71s$ \\
        \bottomrule
	\end{tabular}
	\begin{tablenotes}
        \scriptsize
        \item[] Note that $N_{\textrm{vocab}}$ is the vocabulary size of lowercase words, $\bar{L}_{video}$ denotes the average length of videos in seconds, $\bar{L}_{query}$ denotes the average number of words in sentence query, $\bar{L}_{moment}$ is the average length of temporal moments in seconds, and $\Delta_{moment}$ is the standard deviation of temporal moment length in seconds.
    \end{tablenotes}
    \end{threeparttable}
	\label{tab-data}
\end{table*}

\subsubsection{Split: Clip Segment Level Detection}
As illustrated in Figure~\ref{fig_split_ideal}, given a long video, VSLNet-L splits it into $K$ clip segments: 
\begin{equation}
    \bm{V}=[\bm{C}_k]_{k=1}^{K}~~\text{and}~~ \bm{C}_k=[\bm{v}_i]_{i=(k-1)\times l}^{k\times l}
\label{eq:video_split}
\end{equation}
where $l$ is the length of each clip segment $\bm{C}_k$, \ie $K\times l=n$. Note that, in our implementation, we perform video split at visual feature level, instead of the untrimmed video itself for computational efficiency. Specifically, we split the features $\bm{V}=[\bm{v}_i]_{i=1}^{n}$ obtained from the pre-trained 3D ConvNet (see Section~\ref{ssec:formulaion}), and use the feature vector $\bm{V}$ in Eq.~\ref{eq:video_split} accordingly.

Each clip segment $\bm{C}_k$ is then processed by feature encoder and CQA, to learn query-attended representations $\bm{C}_{k}^{q} =[\bm{c}_{i}^{q}]_{i=(k-1)\times l}^{k\times l}\in\mathbb{R}^{l\times d}$, as shown in Figure~\ref{fig_architecture}(b). Thus, $\bm{C}_{k}^{q}$ encodes the cross-modal features between clip segment $k$ and query. Then, a Nil Prediction Module (NPM) is introduced in VSLNet-L, to predict whether a clip segment contains or partially contains the temporal moment that corresponds to the text query, as shown in Figure~\ref{fig_architecture}(b). Next, we detail the structure of the NPM following the illustration in Figure~\ref{fig_npm}.

For each clip segment, its query-attended features $\bm{C}_{k}^{q}$ are first encoded by a feature encoder as:
\begin{equation}
    \bm{\widehat{C}}_{k}^{q} = \mathtt{FeatureEncoder}_{\text{NPM}}(\bm{C}_{k}^{q})
\end{equation}
The self-attention mechanism~\cite{Bahdanau2015NeuralMT} is adopted to encode $\bm{\widehat{C}}_{k}^{q}$ into clip representation $\bm{h}_{C_k}^{q}$, and the nil-score is computed as:\begin{equation}
    \begin{aligned}
        \bm{\alpha}_{k} & =\mathtt{SoftMax}(\mathtt{Conv1D}(\bm{\widehat{C}}_{k}^{q})) \\
        \bm{h}_{C_k}^{q} & = \sum\nolimits_{i=1}^{l}\alpha_{k,i}\cdot\bm{\hat{c}}_{(k-1)\times l+i}^{q} \\
        \mathcal{S}_{\text{nil}}^{k} & = \sigma(\mathtt{FFN}(\bm{h}_{C_k}^{q}))
    \end{aligned}
\end{equation}
where $\bm{\alpha}_k\in\mathbb{R}^{l_c}$ and $\bm{h}_{C_k}^{q}\in\mathbb{R}^{d}$. $\mathcal{S}_{\text{nil}}^{k}\in\mathbb{R}$ is a scalar, which indicates the confidence of clip segment $k$ containing the ground truth moment. The loss of NPM is formulated as:
\begin{equation}
    \mathcal{L}_{\text{NPM}}=f_{\text{CE}}(\mathcal{S}_{\text{nil}}, Y_{\text{nil}})
\end{equation}
$Y_{\text{nil}}$ is a $0$-$1$ sequence provided during training. A clip segment is positive (label $1$) \emph{iff} it overlaps with ground truth moment, illustrated in Figure~\ref{fig_split_demo}. Clip segments that do not contain ground truth moment are negative (label $0$). 
After computing the nil-scores of all clip segments, \ie $\mathcal{S}_{\text{nil}}=[\mathcal{S}_{\text{nil}}^{1},\dots,\mathcal{S}_{\text{nil}}^{K}]\in\mathbb{R}^{K}$, we normalize the scores. The output for clip segment $k$ is:
\begin{equation}
    \mathcal{\bar{S}}_{\text{nil}} = \frac{\mathcal{S}_{\text{nil}}}{\max(\mathcal{S}_{\text{nil}})},
    \;\;\;\;
    \bm{\bar{C}}_{k}^{q} = \mathcal{\bar{S}}_{\text{nil}}^{k}\times\bm{\widehat{C}}_{k}^{q}
\end{equation}
where $\mathcal{\bar{S}}_{\text{nil}}$ is the normalized nil-score and $\bm{\bar{C}}_{k}^{q}$ is the re-weighted representations of clip segment $k$. 
The NPM highlights the clip segments that contain the target moment, and suppresses other segments. Then the subsequent modules could localize the result by focusing more on the highlighted segments, which is equivalent to narrowing down the searching scope from a long video to a short segment of it.

\subsubsection{Concat: Video Level Localization}
With the clip segments processed separately in the previous step, we now concatenate the representations of all clip segments, for two reasons. First, a single clip segment may not fully cover the target moment. Second, even if a segment is predicted to be negative (or low confidence), it might provide useful contextual information for localizing the target moment. 
\begin{equation}
    \bm{\bar{C}}_{V}^{q}=[\bm{\bar{C}}_{1}^{q} \parallel \bm{\bar{C}}_{2}^{q} \parallel \dots \parallel \bm{\bar{C}}_{K}^{q}]
\end{equation}
where $\parallel$ denotes the concatenation operator and $\bm{\bar{C}}_{V}^{q}\in\mathbb{R}^{n\times d}$. Accordingly, the input feature $\bm{V}^{q}$ of QGH is replaced by $\bm{\bar{C}}_{V}^{q}$ in VSLNet-L to compute $\mathcal{L}_{\text{QGH}}$ and $\mathcal{L}_{\text{span}}$. The combined training loss for VSLNet-L is:
\begin{equation}
    \mathcal{L} = \mathcal{L}_{\text{span}} + \mathcal{L}_{\text{QGH}} + \mathcal{L}_{\text{NPM}}
\end{equation}

\subsubsection{Multi-scale Split-and-Concat}\label{sssec:multiscale}
Unlike a text document, there are no paragraphs as semantic units in a video. Any split in the video may break important contextual information to different clip segments. Although all clip segments will be concatenated again for video level localization, each clip segment is processed separately, and the contextual information between two segments may not be well captured.

To address this issue, we propose a multi-scale split mechanism, to split the video at different segment lengths, \ie different $K$ (see Figure~\ref{fig_split_multi_scale}). Suppose we use $N_{s}$ different scales, and for each scale we have:
\begin{equation}
    K_i\times l_i=n,\;\;\;\forall\;i=1,2,\dots,N_s
\end{equation}
where $K_i$ and $l_i$ denote the number of clip segments and clip segment length for the $i$-th scale, respectively. Through a multi-scale split, contextual information is well captured. Meanwhile, a multi-scale split also provides variation in training samples for the same video and query pair input. The target moment may be located in multiple different clip segments, and its contexts are also different. Note that the same training process as the single-scale split-and-concat applies, except that the clip segments of each scale are fed separately into VSLNet-L, to optimize the objective.

Consequently, we derive $N_s$ predicted moments for a given video and language query pair, because the clip segments at each scale would lead to a pair of start/end boundaries for a predicted moment. During inference, we adopt two simple candidate selection strategies to derive the final target moment. VSLNet-L-$P_m$ strategy selects the candidate with the highest joint boundary probability, $P_{m}=\max\{P_{\text{span}}^{i}\}_{i=1}^{N_s}$, where $P_{\text{span}}^{i}=P_{s}^{i}(\hat{a}^{s})P_{e}^{i}(\hat{a}^{e})$ is the maximal joint boundary probability of the moment generated by $i$-th scale using Eq.~\ref{eq:infer}. VSLNet-L-$U$ strategy selects two moments with the largest overlap from $N_s$ candidates, and computes their union as the final result.

\section{Experiments}\label{sec:exp}

\subsection{Datasets}
We conduct experiments on three benchmark datasets: Charades-STA~\cite{Gao2017TALLTA}, ActivityNet Captions~\cite{krishna2017dense}, and TACoS~\cite{regneri2013grounding}, as summarized in Table~\ref{tab-data}.
\textbf{Charades-STA} is prepared by Gao \etal~\cite{Gao2017TALLTA} based on Charades dataset~\cite{sigurdsson2016hollywood}, with the videos of daily indoor activities. There are $12,408$ and $3,720$ moment annotations for training and test, respectively.
\textbf{ActivityNet Captions (ANetCap)} contains about $20$k videos taken from ActivityNet~\cite{heilbron2015activitynet}. We follow the setup in Yuan \etal~\cite{Yuan2019ToFW}, with $37,421$ moment annotations for training, and $17,505$ annotations for test.
\textbf{TACoS} is selected from MPII Cooking Composite Activities dataset~\cite{Rohrbach2012SDA}. There are two versions of TACoS available. TACoS$_{\text{org}}$ used in~\cite{Gao2017TALLTA} has $10,146$, $4,589$ and $4,083$ annotations for training, validation and test. TACoS$_{\text{tan}}$ is from 2D-TAN~\cite{zhang2020learning}, which contains $9,790$, $4,436$ and $4,001$ annotations in train, validation and test sets.

We conduct experiments on all the benchmark datasets for VSLBase and VSLNet. For VSLNet-L, we evaluate it on ActivityNet Captions and both versions of TACoS. Charades-STA is not used for VSLNet-L due to its short video length.

\begin{table}[t]
   \small
    \caption{\small Results ($\%$) of ``$\textrm{Rank@}1, \textrm{IoU}=\mu$'' and ``mIoU'' compared with the state-of-the-art on Charades-STA.}
    \vspace{-0.3cm}
	\centering
	\begin{tabular}{l c c c c}
		\toprule
		\multirow{2}{*}{Model} & \multicolumn{3}{c}{$\text{Rank@}1, \text{IoU}=\mu$} & \multirow{2}{*}{mIoU} \\
        & $\mu=0.3$ & $\mu=0.5$ & $\mu=0.7$ & \\
        \midrule
        \multicolumn{5}{c}{3D ConvNet without fine-tuning as visual feature extractor} \\
        \midrule
        CTRL~\cite{Gao2017TALLTA}             & - & 23.63 & 8.89  & - \\
        ACL~\cite{ge2019mac}                  & - & 30.48 & 12.20 & - \\
        QSPN~\cite{Xu2019MultilevelLA}        & 54.70 & 35.60 & 15.80 & - \\
        SAP~\cite{chen2019semantic}           & - & 27.42 & 13.36 & - \\
        SM-RL~\cite{Wang2019LanguageDrivenTA} & - & 24.36 & 11.17 & - \\
        RWM-RL~\cite{he2019Readwa}            & - & 36.70 & - & - \\
        MAN~\cite{zhang2019man}               & - & \underline{46.53} & 22.72 & - \\
        DEBUG~\cite{lu2019debug}              & 54.95 & 37.39 & 17.69 & 36.34 \\
        TSP-PRL~\cite{Wu2020TreeStructuredPB} & - & 37.39 & 17.69 & 37.22 \\
        2D-TAN~\cite{zhang2020learning}       & - & 39.81 & 23.31 & - \\
        CBP~\cite{Wang2020TemporallyGL}       & - & 36.80 & 18.87 & - \\
        GDP~\cite{chen2020rethinking}         & 54.54 & 39.47 & 18.49 & - \\
        \midrule
        VSLBase                               & \underline{61.72} & 40.97 & \underline{24.14} & \underline{42.11} \\
        VSLNet                                & \textbf{64.30} & \textbf{47.31} & \textbf{30.19} & \textbf{45.15} \\
        \midrule
        \multicolumn{5}{c}{3D ConvNet with fine-tuning on Charades dataset} \\
        \midrule
        ExCL~\cite{ghosh2019excl}             & 65.10 & 44.10 & 23.30 & - \\
        SCDM~\cite{yuan2019semantic}          & - & \textbf{54.44} & \underline{33.43} & - \\
        DRN~\cite{zeng2020dense}              & - & $53.09$ & $31.75$ & - \\
        \midrule
        VSLBase                               & \underline{68.06} & $50.23$ & $30.16$ & \underline{47.15} \\
        VSLNet                                & \textbf{70.46} & \underline{54.19} & \textbf{35.22} & \textbf{50.02} \\
        \bottomrule
	\end{tabular}
	\label{tab-sota-charades}
\end{table}

\begin{table}[t]
    \small
	\caption{\small Results ($\%$) of ``$\text{Rank@}1, \text{IoU}=\mu$'' and ``mIoU'' compared with the state-of-the-art on TACoS.}
	\vspace{-0.3cm}
	\centering
	\setlength{\tabcolsep}{3.6pt}
	\begin{threeparttable}
	\begin{tabular}{c l c c c c}
		\toprule
        \multirow{2}{*}{Dataset} & \multirow{2}{*}{Model} & \multicolumn{3}{c}{$\text{Rank@}1, \text{IoU}=\mu$} & \multirow{2}{*}{mIoU} \\
        & & $\mu=0.3$ & $\mu=0.5$ & $\mu=0.7$ & \\
        \midrule
        \multirow{18}{*}{\tabincell{l}{\rotatebox{270}{TACoS$_{\text{org}}$}}} 
        & CTRL~\cite{Gao2017TALLTA}             & 18.32 & 13.30 & -  & - \\
        & TGN~\cite{chen2018temporally}         & 21.77 & 18.90 & - & - \\
        & ACL~\cite{ge2019mac}                  & 24.17 & 20.01 & - & - \\
        & ACRN~\cite{Liu2018AMR}                & 19.52 & 14.62 & - & - \\
        & ABLR~\cite{Yuan2019ToFW}              & 19.50 & 9.40 & - & 13.40 \\
        & SM-RL~\cite{Wang2019LanguageDrivenTA} & 20.25 & 15.95 & - & - \\
        & DEBUG~\cite{lu2019debug}              & 23.45 & 11.72 & - & 16.03 \\
        & SCDM~\cite{yuan2019semantic}          & 26.11 & 21.17 & - & - \\
        & GDP~\cite{chen2020rethinking}         & 24.14 & 13.90 & - & 16.18 \\
        & CBP~\cite{Wang2020TemporallyGL}       & 27.31 & 24.79 & 19.10 & 21.59 \\
        & DRN~\cite{zeng2020dense}              & - & 23.17 & - & - \\
        \cmidrule{2-6}
        & VSLNet$^{(S)}$                        & 29.21 & 24.37 & 19.37 & 23.48 \\
        & VSLNet$^{(L)}$                        & 29.78 & 24.71 & 19.64 & 23.96 \\
        \cmidrule{2-6}
        & VSLNet-L-$P_m$$^{(S)}$                & \textbf{31.94} & \underline{26.72} & \textbf{22.36} & \underline{25.71} \\
        & VSLNet-L-$U$$^{(S)}$                  & \underline{31.69} & \textbf{26.79} & \underline{22.02} & \textbf{25.78} \\
        \cmidrule{2-6}
        & VSLNet-L-$P_m$$^{(L)}$                & \textbf{32.04} & \textbf{27.92} & \textbf{23.28} & \textbf{26.40} \\
        & VSLNet-L-$U$$^{(L)}$                  & \underline{31.86} & \underline{27.64} & \underline{22.72} & \underline{26.25} \\
        \midrule
        \multirow{8}{*}{\tabincell{c}{\rotatebox{270}{TACoS$_{\text{tan}}$}}} 
        & 2D-TAN Pool~\cite{zhang2020learning}  & 37.29 & 25.32 & - & - \\
        & 2D-TAN Conv~\cite{zhang2020learning}  & 35.22 & 25.19 & - & - \\
        \cmidrule{2-6}
        & VSLNet$^{(S)}$                        & 42.66 & 32.72 & 23.12 & 33.07 \\
        & VSLNet$^{(L)}$                        & 41.42 & 30.67 & 22.32 & 31.92 \\
        \cmidrule{2-6}
        & VSLNet-L-$P_m$$^{(S)}$                & \textbf{47.66} & \textbf{36.15} & \textbf{26.19} & \textbf{36.24} \\
        & VSLNet-L-$U$$^{(S)}$                  & \textbf{47.66} & \underline{36.12} & \underline{25.87} & \underline{35.98} \\
        \cmidrule{2-6}
        & VSLNet-L-$P_m$$^{(L)}$                & \textbf{47.11} & \textbf{36.34} & \textbf{26.42} & \textbf{36.61} \\
        & VSLNet-L-$U$$^{(L)}$                  & \underline{46.44} & \underline{35.74} & \underline{26.19} & \underline{36.05} \\
        \bottomrule
	\end{tabular}
	\begin{tablenotes}
        \scriptsize
        \item[] $(S)$ denotes $n=300$ and $(L)$ represents $n=600$.
    \end{tablenotes}
    \end{threeparttable}
	\label{tab-sota-tacos}
\end{table}

\begin{table}[t]
    \small
    \caption{\small Results ($\%$) of ``$\text{Rank@}1, \text{IoU}=\mu$'' and ``mIoU'' compared with state-of-the-arts on ActivityNet Captions.}
    \vspace{-0.3cm}
	\centering
	\begin{threeparttable}
	\begin{tabular}{l c c c c}
		\toprule
        \multirow{2}{*}{Model} & \multicolumn{3}{c}{$\text{Rank@}1, \text{IoU}=\mu$} & \multirow{2}{*}{mIoU} \\
        & $\mu=0.3$ & $\mu=0.5$ & $\mu=0.7$ & \\
        \midrule
        TGN~\cite{chen2018temporally}         & 45.51 & 28.47 & -     & -     \\
        ACRN~\cite{Liu2018AMR}                & 49.70 & 31.67 & 11.25 & -     \\
        ABLR~\cite{Yuan2019ToFW}              & 55.67 & 36.79 & -     & 36.99 \\
        RWM-RL~\cite{he2019Readwa}            & -     & 36.90 & -     & -     \\
        QSPN~\cite{Xu2019MultilevelLA}        & 45.30 & 27.70 & 13.60 & -     \\
        DEBUG~\cite{lu2019debug}              & 55.91 & 39.72 & -     & 39.51 \\
        SCDM~\cite{yuan2019semantic}          & 54.80 & 36.75 & 19.86 & -     \\
        TSP-PRL~\cite{Wu2020TreeStructuredPB} & 56.08 & 38.76 & -     & 39.21 \\
        GDP~\cite{chen2020rethinking}         & 56.17 & 39.27 & -     & 39.80 \\
        CBP~\cite{Wang2020TemporallyGL}       & 54.30 & 35.76 & 17.80 & -     \\
        DRN~\cite{zeng2020dense}              & -     & \textbf{45.45} & 24.36 & -     \\
        LGI~\cite{mun2020local}               & 58.52 & 41.51 & 23.07 & -     \\
        2D-TAN Pool~\cite{zhang2020learning}  & 59.45 & \underline{44.51} & 26.54 & -     \\
        2D-TAN Conv~\cite{zhang2020learning}  & 58.75 & 44.05 & \underline{27.38} & -     \\
        \midrule
        VSLBase*                              & 58.18 & 39.52 & 23.21 & 40.56 \\
        VSLNet*                               & \textbf{63.16} & 43.22 & 26.16 & 43.19 \\
        VSLNet                                & 61.61 & 43.78 & 26.54 & 43.22 \\
        VSLNet-L-$P_m$                        & 62.18 & 43.69 & 27.22 & \underline{43.67} \\
        VSLNet-L-$U$                          & \underline{62.35} & 43.86 & \textbf{27.51} & \textbf{44.06} \\
        \bottomrule
	\end{tabular}
	\begin{tablenotes}
        \scriptsize
        \item[] * denotes that the maximal visual sequence length $n$ of VSLBase and VSLNet is set as $128$, which is consistent with~\cite{zhang2020vslnet}.
    \end{tablenotes}
    \end{threeparttable}
	\label{tab-sota-activitynet}
\end{table}

\subsection{Experimental Settings}

\subsubsection{Evaluation Metrics}
We adopt ``$\textrm{Rank@}n, \textrm{IoU}=\mu$'' and ``mIoU'' as the evaluation metrics, following~\cite{Gao2017TALLTA,Liu2018AMR,Yuan2019ToFW}. The ``$\textrm{Rank@}n, \textrm{IoU}=\mu$'' denotes the percentage of language queries having at least one result whose Intersection over Union (IoU) with ground truth is larger than $\mu$ in top-\textit{n} retrieved moments. ``mIoU'' is the average IoU over all testing samples. In our experiments, we use $n=1$ and $\mu\in\{0.3, 0.5, 0.7\}$.

\subsubsection{Implementation Details}
For a text query $Q$, we lowercase all its words and initialize them with $300$d GloVe~\cite{pennington2014glove} embeddings. The word embeddings are fixed during training. For the untrimmed video $V$, we extract its visual features using 3D ConvNet pre-trained on Kinetics dataset~\cite{Carreira2017QuoVA}. We set the maximal feature length $n$ as $128$ for Charades-STA, \ie the extracted visual feature sequence of a video will be uniformly downsampled if its length $>n$, or zero-padding otherwise. The $n$ of ANetCap is set to $300$, while two maximal feature lengths, $n\in\{300, 600\}$, are used for evaluation on TACoS. When evaluating VSLNet-L, the visual features are split into multiple clip segments using different scales, we use $l=\{60, 75, 100, 150\}$ (\ie $K=\{5, 4, 3, 2\}$) for $n=300$, and $l=\{100, 120, 150, 200\}$ (\ie $K=\{6, 5, 4, 3\}$) for $n=600$. We set the dimension of all the hidden layers in the model as $128$; the kernel size of the convolution layer is $7$; the head size of multi-head attention is $8$. All the models are trained for $100$ epochs with a batch size of $16$ and an early stopping strategy for all datasets. Adam~\cite{Kingma2015AdamAM} is used as the optimizer, with a learning rate of $0.0001$, linear decay of learning rate and gradient clipping of $1.0$. Dropout~\cite{srivastava2014dropout} of $0.2$ is applied to prevent overfitting. All experiments are conducted on dual NVIDIA GeForce RTX 2080Ti GPUs workstation.

\begin{figure}[t]
    \centering
	\includegraphics[trim={0cm 0cm 0cm 0cm},clip,width=0.4\textwidth]{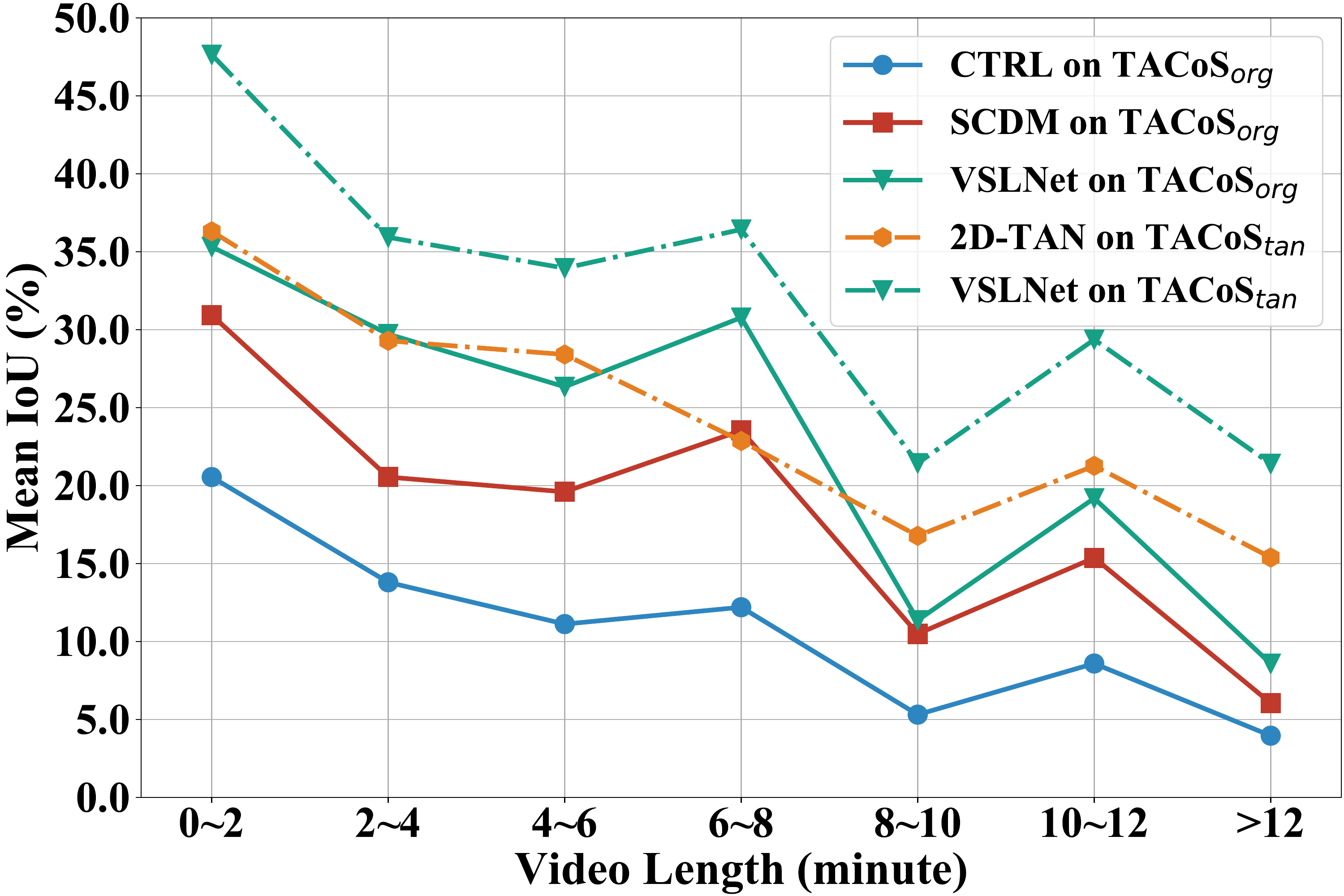}
	\vspace{-0.3cm}
	\caption{\small The Mean IoU (\%) performance of CTRL~\cite{Gao2017TALLTA}, SCDM~\cite{yuan2019semantic}, 2D-TAN Pool~\cite{zhang2020learning} and VSLNet on the TACoS dataset.}
	\label{fig_tacos_result_diff_length}
\end{figure}

\begin{figure}[t]
    \centering
	\includegraphics[trim={0cm 0cm 0cm 0cm},clip,width=0.4\textwidth]{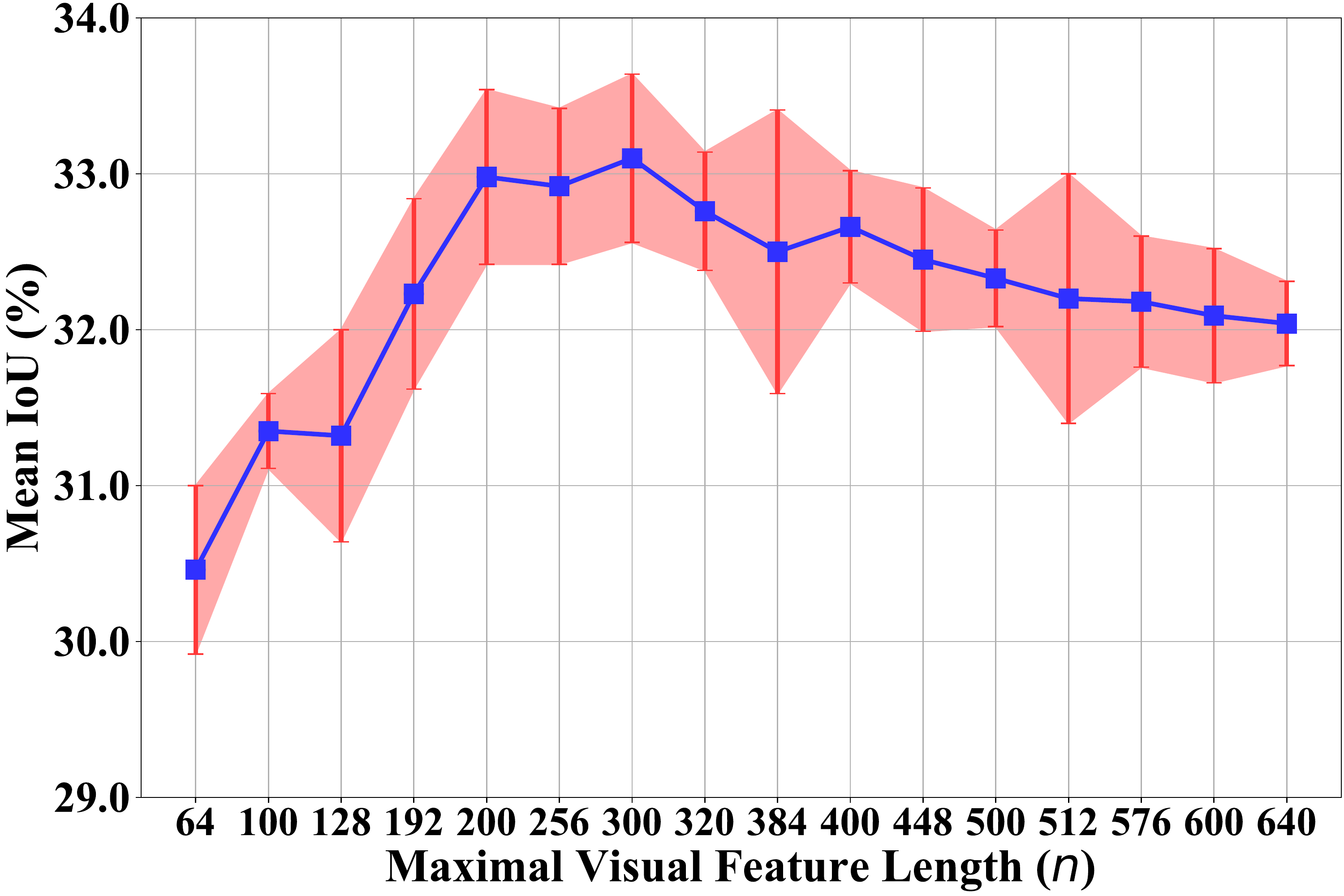}
	\vspace{-0.3cm}
	\caption{\small Mean IoU ($\%$) results of VSLNet on the TACoS$_{\text{tan}}$ dataset under different maximal visual representation lengths $n$.}
	\label{fig_mean_iou_n}
\end{figure}

\begin{figure}[t]
    \centering
	\includegraphics[trim={0cm 0cm 0cm 0cm},clip,width=0.4\textwidth]{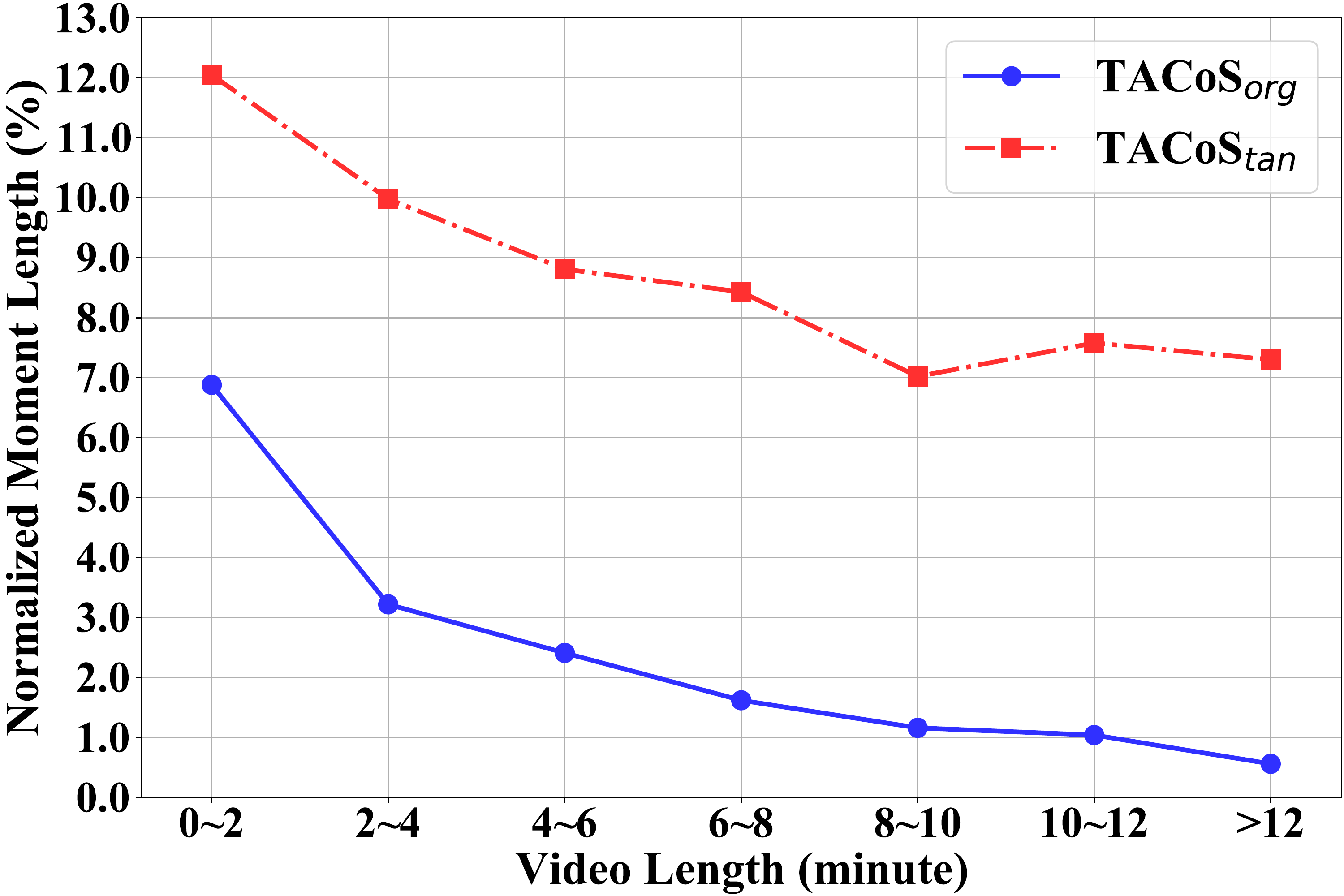}
	\vspace{-0.3cm}
	\caption{\small Statistic of normalized moment lengths in videos for both TACoS$_{\text{org}}$ and TACoS$_{\text{tan}}$.}
	\label{fig_tacos_moment_ratio}
\end{figure}

\begin{figure}[t]
    \centering
	\includegraphics[trim={0cm 0cm 0cm 0cm},clip,width=0.4\textwidth]{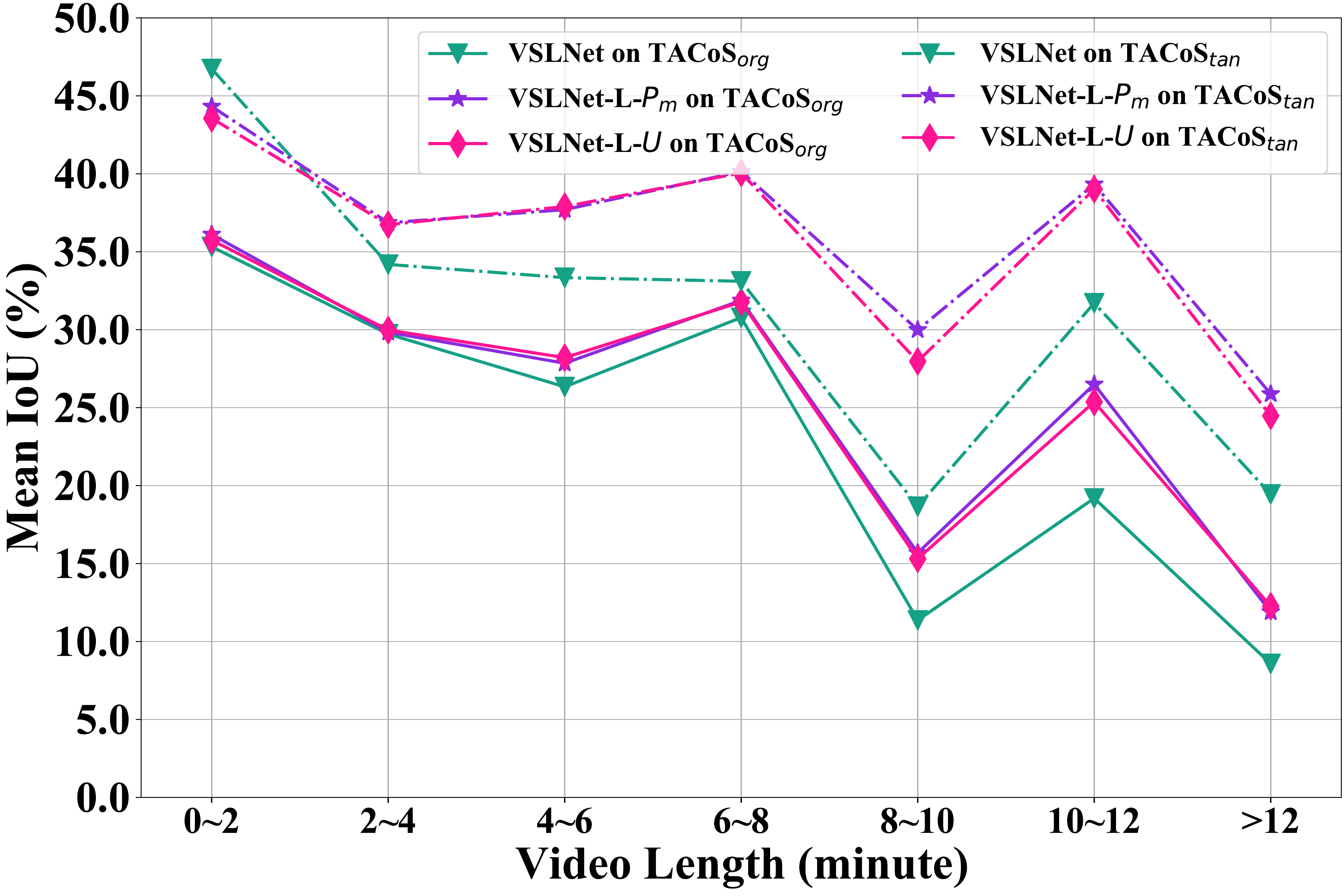}
	\vspace{-0.3cm}
	\caption{\small Visualization of performance improvement of VSLNet-L on different video lengths compared to VSLNet on TACoS.}
	\label{fig_tacos_diff_length}
\end{figure}

\subsection{Comparison with State-of-the-Arts}\label{subsec:sota}
We compare the proposed methods with the following state-of-the-arts: CTRL~\cite{Gao2017TALLTA}, ACRN~\cite{Liu2018AMR}, TGN~\cite{chen2018temporally}, ACL~\cite{ge2019mac}, DEBUG~\cite{lu2019debug}, SAP~\cite{chen2019semantic}, QSPN~\cite{Xu2019MultilevelLA}, SM-RL~\cite{Wang2019LanguageDrivenTA}, RWM-RL~\cite{he2019Readwa}, ABLR~\cite{Yuan2019ToFW}, MAN~\cite{zhang2019man}, SCDM~\cite{yuan2019semantic}, TSP-PRL~\cite{Wu2020TreeStructuredPB}, 2D-TAN~\cite{zhang2020learning}, GDP~\cite{chen2020rethinking}, ExCL~\cite{ghosh2019excl}, CBP~\cite{Wang2020TemporallyGL}, LGI~\cite{mun2020local} and DRN~\cite{zeng2020dense}. Scores of the benchmark methods in all result tables are those reported in the corresponding papers. The best results are in \textbf{bold} and the second bests are \underline{underlined}.

Among the proposed methods, VSLBase is a direct implementation of span-based QA framework on the NLVL task;  VSLNet is the extension of VSLBase with QGH;  VSLNet-L is a further extension of VSLNet with multi-scale split-concat strategy, designed to handle long videos more effectively. In the following, we show that VSLBase is comparable to existing baselines on NLVL tasks, while VSLNet surpasses VSLBase and all existing baselines, and achieves state-of-the-art results. We then show that VSLNet-L well addresses the issue of performance degradation on NLVL for long videos, compared to VSLNet.

The results on Charades-STA are reported in Table~\ref{tab-sota-charades}. It is observed that VSLNet significantly outperforms most of the baselines by a large margin over all metrics. In addition, it is worth noting that the performance improvements of VSLNet are more significant under larger values of IoU. For instance, VSLNet achieves $7.47\%$ improvement in $\textrm{IoU}=0.7$ versus $0.78\%$ in $\textrm{IoU}=0.5$, compared to MAN. Without query-guided highlighting, VSLBase outperforms all compared baselines over $\textrm{IoU}=0.7$, which shows adopting the span-based QA framework is promising for NLVL. Moreover, VSLNet benefits from visual feature fine-tuning, and achieves state-of-the-art results on this dataset.

Table~\ref{tab-sota-tacos} reports the results on both versions (if available) of TACoS dataset. In general, VSLNet outperforms previous methods over all evaluation metrics. In addition, with the Split-and-Concat mechanism, VSLNet-L further improves the performance, on top of VSLNet. On TACoS$_{\text{org}}$, the results of VSLNet$^{(S)}$ is comparable to that of VSLNet$^{(L)}$, while VSLNet-L$^{(L)}$ surpasses VSLNet-L$^{(S)}$ for both candidate selection strategies. Here, $L$ and $S$ denote the maximal video feature length $600$ and $300$, respectively. Similar observations hold on TACoS$_{\text{tan}}$. These results demonstrate that VSLNet-L is more adept than others at localizing temporal moments in longer videos. Moreover, VSLNet-L-$P_m$ is generally superior to VSLNet-L-$U$ under different $n$ for both versions of the TACoS dataset.

The results on the ActivityNet Captions dataset are summarized in Table~\ref{tab-sota-activitynet}. We observe that VSLBase shows similar performance to or slightly better than most of the baselines, while VSLNet further boosts the performance of VSLBase significantly. Comparing VSLNet with $n=128$ and that with $n=300$, we find that small $n$ leads to better performance on loose metric (\eg $63.16$ versus $61.61$ on $\text{IoU}=0.3$) and large $n$ is beneficial for strict metric (\eg $26.16$ versus $26.54$ on $\text{IoU}=0.7$). Meanwhile, the performance of VSLNet-L is comparable to state-of-the-art methods. It is worth noting that $99\%$ annotations in ActivityNet Captions belong to videos that are shorter than $4$ minutes. As VSLNet-L is designed to address performance degradation on long videos, it is reasonable to observe that VSLNet-L achieves less significant performance improvement on ActivityNet Captions compared to TACoS, \textit{w.r.t.} the state-of-the-arts. Moreover, VSLNet-L-$U$ performs better than VSLNet-L-$P_m$ on ActivityNet Captions, different from the observation on TACoS. This could be due to the different ratios of $\bar{L}_{\text{moment}}$ and  $\bar{L}_{\text{video}}$ in the two datasets (see Table~\ref{tab-data}). The strategy to select longer spans works better on ActivityNet Captions dataset.

\begin{table*}[t]
    \small
    \caption{\small Statistics of videos and annotations \textit{w.r.t.} different video lengths over NLVL datasets.}
	\vspace{-0.3cm}
	\centering
	\setlength{\tabcolsep}{3.6 pt}
	\begin{tabular}{c c c c c c c c c c c}
	    \toprule
	    \multirow{2}{*}{Dataset} & \multirow{2}{*}{Split} & \multirow{2}{*}{\# Videos} & \multirow{2}{*}{\# Annots} & \multicolumn{7}{c}{\# of videos / annotations \textit{w.r.t.} different video lengths} \\
	    & & & & $0\sim2$ min & $2\sim4$ min & $4\sim6$ min & $6\sim8$ min & $8\sim10$ min & $10\sim12$ min & $>12$ min \\
	    \midrule
	    \multirow{3}{*}{TACoS$_{\text{org}}$} & Train & $75$ & $10,146$ & $27$ / $2,847$ & $29$ / $4,015$ & $8$ / $1,284$ & $4$ / $607$ & $3$ / $616$ & $2$ / $328$ & $2$ / $449$ \\
	    & Val & $27$ & $4,589$ & $3$ / $312$ & $5$ / $771$ & $5$ / $887$ & $7$ / $1,275$ & $1$ / $173$ & $4$ / $830$ & $2$ / $341$ \\
	    & Test & $25$ & $4,083$ & $5$ / $578$ & $6$ / $937$ & $3$ / $564$ & $3$ / $447$ & $2$ / $373$ & $3$ / $617$ & $3$ / $567$ \\
	    \midrule
	    \multirow{3}{*}{TACoS$_{\text{tan}}$} & Train & $75$ & $9,790$ & $27$ / $2,769$ & $29$ / $3,840$ & $8$ / $1,227$ & $4$ / $576$ & $3$ / $597$ & $2$ / $336$ & $2$ / $445$ \\
	    & Val & $27$ & $4,436$ & $3$ / $311$ & $6$ / $929$ & $4$ / $639$ & $7$ / $1,225$ & $1$ / $171$ & $4$ / $812$ & $2$ / $349$ \\
	    & Test & $25$ & $4,001$ & $5$ / $594$ & $6$ / $907$ & $3$ / $535$ & $3$ / $428$ & $2$ / $370$ & $3$ / $598$ & $3$ / $569$ \\
	    \midrule
	    \multirow{2}{*}{ANetCap} & Train & $10,009$ & $37,421$ & $5,278$ / $17,806$ & $4,715$ / $19,551$ & $8$ / $28$ & $3$ / $10$ & $4$ / $22$ & $0$ / $0$ & $1$ / $4$ \\
	    & Test & $4,917$ & $17,505$ & $2,516$ / $8,193$ & $2,392$ / $9,274$ & $5$ / $19$ & $2$ / $9$ & $0$ / $0$ & $1$ / $5$ & $1$ / $5$ \\
	    \bottomrule
	\end{tabular}
	\label{tab:stat_video_annot_length}
\end{table*}

\begin{table*}[t]
    \small
	\caption{\small Comparison of mIoU ($\%$) between VSLNet and VSLNet-L on TACoS dataset \textit{w.r.t.} different video lengths.}
	\vspace{-0.3cm}
	\centering
	\begin{threeparttable}
	\begin{tabular}{c l l l l l l l l}
		\toprule
        Dataset & Video Length & $0\sim2$ min & $2\sim4$ min & $4\sim6$ min & $6\sim8$ min & $8\sim10$ min & $10\sim12$ min & $>12$ min \\
        \midrule
        \multirow{7}{*}{\tabincell{c}{\rotatebox{270}{TACoS$_{\text{org}}$}}} & \# Test Annotations & $578$ & $937$ & $564$ & $447$ & $373$ & $617$ & $567$ \\
        \cmidrule{2-9}
        & VSLNet$^{(S)}$ & 38.93 & 27.68 & 21.56 & 28.95 & 12.12 & 20.59 & 9.01 \\
        & VSLNet-L-$P_m$$^{(S)}$ & \textbf{39.91} \red{+0.98} & \underline{28.11} \red{+0.43} & \underline{24.63} \red{+3.07} & \textbf{30.45} \red{+1.50} & \underline{15.26} \red{+3.14} & \underline{24.56} \red{+3.97} & \textbf{12.75} \red{+3.74} \\
        & VSLNet-L-$U$$^{(S)}$ & \underline{39.54} \red{+0.61} & \textbf{28.69} \red{+1.01} & \textbf{24.85} \red{+3.29} & \underline{30.37} \red{+1.42} & \textbf{15.27} \red{+3.15} & \textbf{24.87} \red{+4.28} & \underline{12.15} \red{+3.14} \\
        \cmidrule{2-9}
        & VSLNet$^{(L)}$ & 35.32 & 29.72 & 26.34 & 30.78 & 11.37 & 19.19 & 8.58 \\
        & VSLNet-L-$P_m$$^{(L)}$ & \textbf{36.11} \red{+0.79} & \underline{29.84} \red{+0.12} & \underline{27.85} \red{+1.51} & \textbf{31.88} \red{+1.10} & \textbf{15.68} \red{+4.31} & \textbf{26.50} \red{+7.31} & \underline{11.90} \red{+3.32} \\
        & VSLNet-L-$U$$^{(L)}$ & \underline{35.75} \red{+0.43} & \textbf{29.98} \red{+0.26} & \textbf{28.21} \red{+1.87} & \underline{31.78} \red{+1.00} & \underline{15.31} \red{+3.94} & \underline{25.37} \red{+6.18} & \textbf{12.26} \red{+3.68} \\
        \midrule
        \multirow{7}{*}{\tabincell{c}{\rotatebox{270}{TACoS$_{\text{tan}}$}}} & \# Test Annotations & $594$ & $907$ & $535$ & $428$ & $370$ & $598$ & $569$ \\
        \cmidrule{2-9}
        & VSLNet$^{(S)}$ & \textbf{47.64} & 35.93 & 33.95 & 36.43 & 21.44 & 29.37 & 21.40 \\
        & VSLNet-L-$P_m$$^{(S)}$ & \underline{46.12} \green{-1.52} & \textbf{37.76} \red{+1.83} & \underline{36.51} \red{+2.56} & \underline{41.37} \red{+4.94} & \textbf{29.71} \red{+8.27} & \textbf{34.84} \red{+5.47} & \underline{25.08} \red{+3.68} \\
        & VSLNet-L-$U$$^{(S)}$ & 45.91 \green{-1.73} & \underline{37.52} \red{+1.59} & \textbf{36.85} \red{+2.90} & \textbf{41.85} \red{+5.42} & \underline{28.30} \red{+6.86} & \underline{33.84} \red{+4.47} & \textbf{25.20} \red{+3.80} \\
        \cmidrule{2-9}
        & VSLNet$^{(L)}$ & \textbf{46.73} & 34.19 & 33.34 & 33.11 & 18.66 & 31.71 & 19.45 \\
        & VSLNet-L-$P_m$$^{(L)}$ & \underline{44.32} \green{-2.41} & \textbf{36.86} \red{+2.67} & \underline{37.70} \red{+4.36} & \textbf{40.15} \red{+7.04} & \textbf{29.99} \red{+11.33} & \textbf{39.33} \red{+7.62} & \textbf{25.88} \red{+6.43} \\
        & VSLNet-L-$U$$^{(L)}$ & 43.56 \green{-3.17} & \underline{36.74} \red{+2.55} & \textbf{37.90} \red{+4.56} & \underline{40.06} \red{+6.95} & \underline{27.97} \red{+9.31} & \underline{39.01} \red{+7.30} & \underline{24.48} \red{+5.03} \\
        \bottomrule
	\end{tabular}
	\begin{tablenotes}
        \scriptsize
        \item[] $(S)$ denotes $n=300$ and $(L)$ represents $n=600$. Performance \red{gain} and \green{loss} are indicated in different colors.
    \end{tablenotes}
    \end{threeparttable}
	\label{tab:duration_tacos}
\end{table*}

\begin{table}[t]
    \small
    \caption{\small Comparison of mIoU ($\%$) between VSLNet and VSLNet-L on ActivityNet Captions \textit{w.r.t.} different video lengths.}
    \vspace{-0.3cm}
	\centering
	\begin{threeparttable}
	\begin{tabular}{l l l l}
        \toprule
        Video Length & $0\sim2$ min & $2\sim4$ min & $>4$ min \\
        \midrule
        \# Test Samples & $8,193$ & $9,274$ & $38$ \\
        \midrule
        VSLNet & 46.21 & 40.60 & 35.24 \\
        VSLNet-L-$P_m$ & \underline{46.59} \red{+0.38} & \underline{41.11} \red{+0.51} & \underline{38.40} \red{+3.16} \\
        VSLNet-L-$U$ & \textbf{47.03} \red{+0.82} & \textbf{41.46} \red{+0.86} & \textbf{39.63} \red{+4.39} \\
        \bottomrule
	\end{tabular}
	\begin{tablenotes}
        \scriptsize
        \item[] Performance \red{gain} and \green{loss} are indicated in different colors.
    \end{tablenotes}
    \end{threeparttable}
	\label{tab:duration_activitynet}
\end{table}

\begin{table}[t]
    \small
    \caption{\small Comparison between models with alternative modules in VSLBase on Charades-STA.}
    \vspace{-0.3cm}
	\centering
	\begin{tabular}{l c c c c}
		\toprule
        \multirow{2}{*}{Module} & \multicolumn{3}{c}{$\text{Rank@}1, \text{IoU}=\mu$} & \multirow{2}{*}{mIoU} \\
        & $\mu=0.3$ & $\mu=0.5$ & $\mu=0.7$ & \\
        \midrule
        BiLSTM + CAT & 61.18 & 43.04 & 26.42 & 42.83 \\
        CMF + CAT    & 63.49 & 44.87 & 27.07 & 44.01 \\
        BiLSTM + CQA & 65.08 & 46.94 & 28.55 & 45.18 \\
        CMF + CQA    & 68.06 & 50.23 & 30.16 & 47.15 \\
        \bottomrule
	\end{tabular}
	\label{tab-component}
\end{table}

\begin{table}[t]
    \small
    \caption{\small Performance gains ($\%$) of different modules over ``$\textrm{Rank@}1, \textrm{IoU}=0.7$'' on Charades-STA.}
    \vspace{-0.3cm}
	\centering
	\begin{tabular}{l c c c}
		\toprule
        Module & CAT & CQA & $\Delta$ \\
        \midrule
        BiLSTM   & 26.42 & 28.55 & +2.13 \\
        CMF      & 27.07 & 30.16 & +3.09 \\
        $\Delta$ & +0.65 & +1.61 & - \\
        \bottomrule
	\end{tabular}
	\label{tab-component-delta}
\end{table}

\begin{figure}[t]
    \centering
	\includegraphics[width=0.48\textwidth]{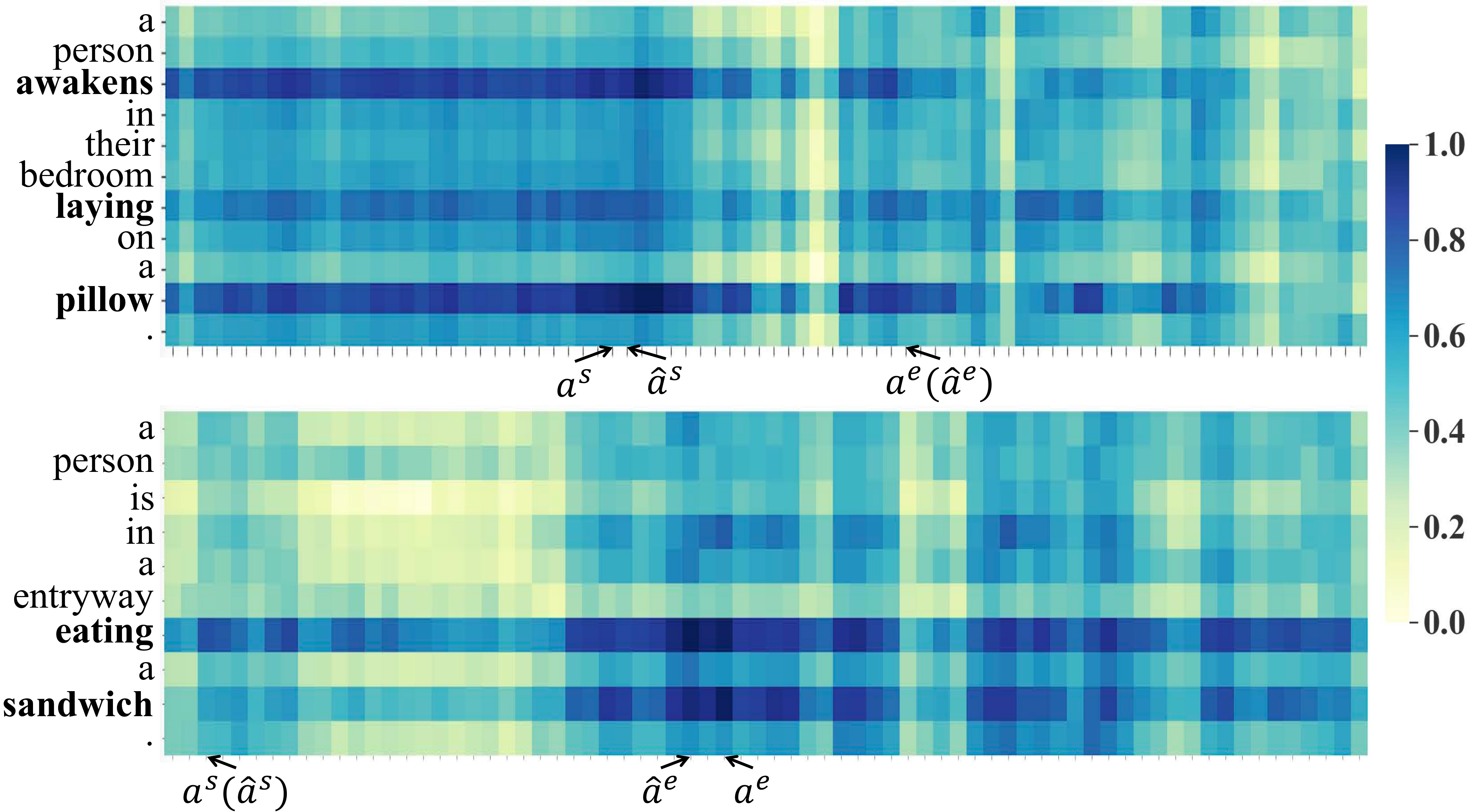}
	\vspace{-0.3cm}
	\caption{\small Similarity scores, $\mathcal{S}$, between visual and language features in the context-query attention.  $a^s/a^e$ denote the start/end boundaries of ground truth video moment, $\hat{a}^{s}/\hat{a}^{e}$ denote the start/end boundaries of predicted target moment.}
	\label{fig_attn_score}
\end{figure}

\begin{figure}[t]
    \centering
	\subfigure[\small $\textrm{Rank@}1, \textrm{IoU}=0.3$]
	{\label{fig_mask_ratio_iou3}	\includegraphics[trim={0.3cm 0.2cm 0.3cm 0.3cm},clip,width=0.23\textwidth]{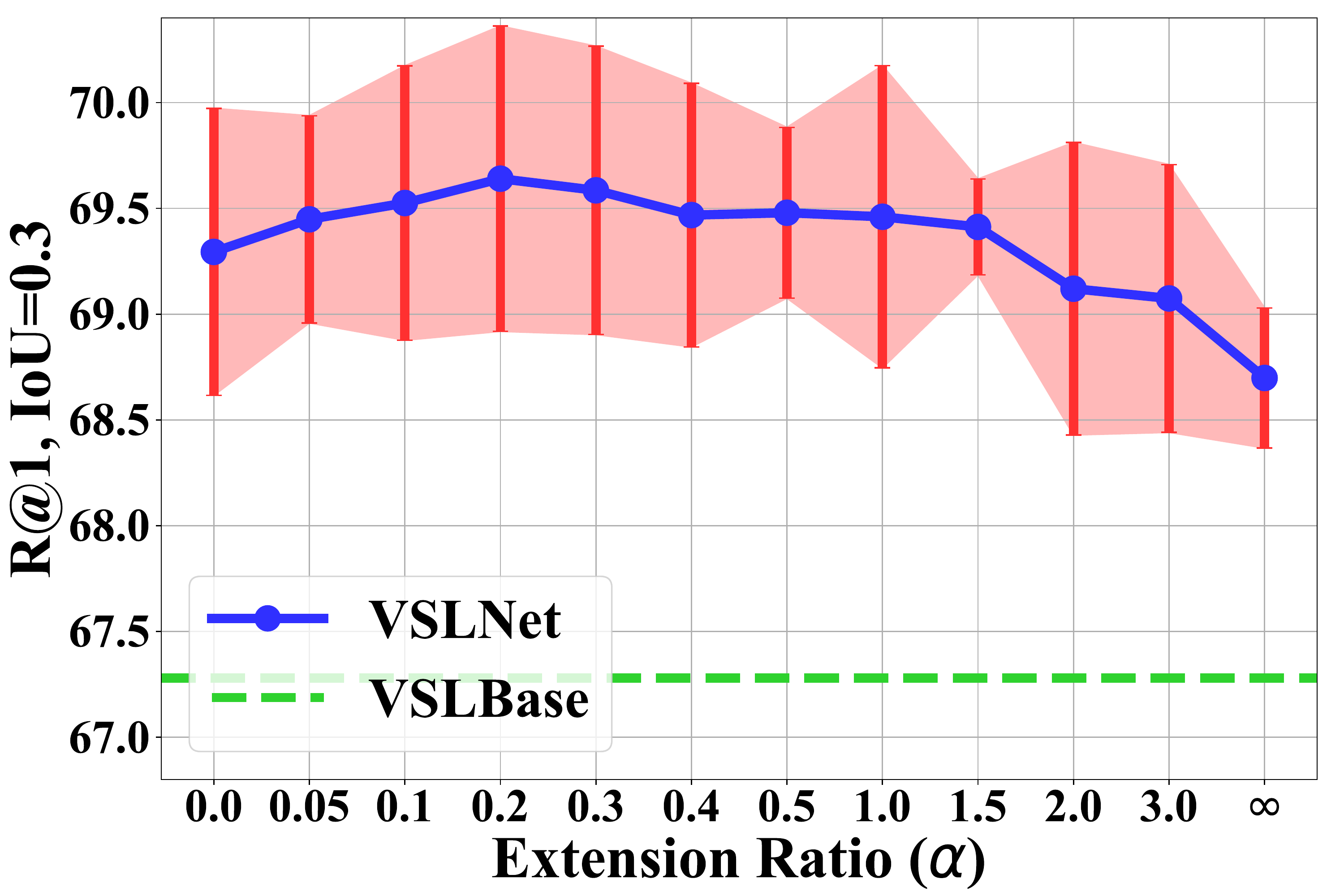}}
	\subfigure[\small $\textrm{Rank@}1, \textrm{IoU}=0.5$]
	{\label{fig_mask_ratio_iou5}	\includegraphics[trim={0.3cm 0.2cm 0.3cm 0.3cm},clip, width=0.23\textwidth]{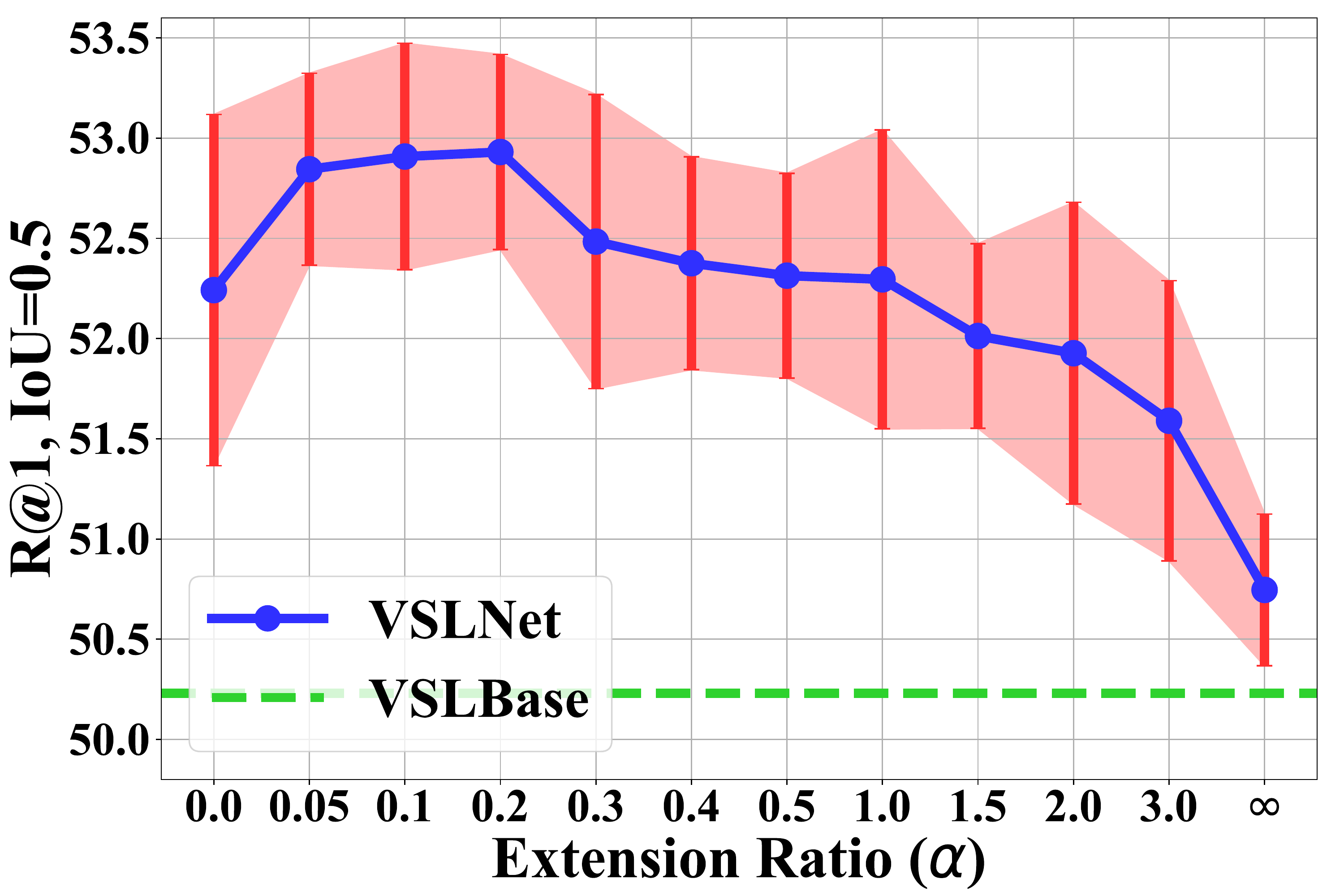}}
	\subfigure[\small $\textrm{Rank@}1, \textrm{IoU}=0.7$]
	{\label{fig_mask_ratio_iou7}	\includegraphics[trim={0.3cm 0.2cm 0.3cm 0.3cm},clip, width=0.23\textwidth]{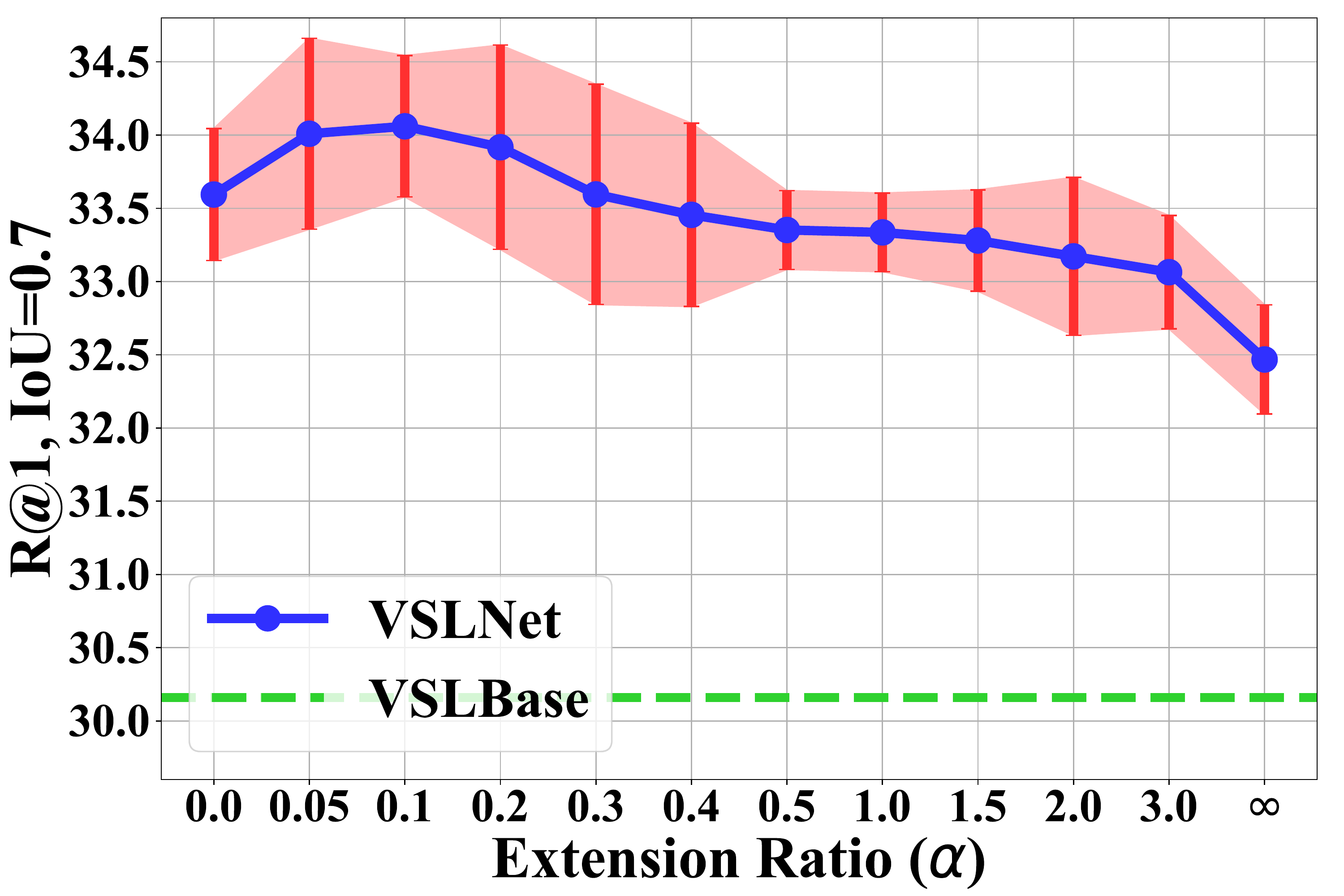}}
	\subfigure[\small mIoU]
	{\label{fig_mask_ratio_miou}	\includegraphics[trim={0.3cm 0.2cm 0.3cm 0.3cm},clip, width=0.23\textwidth]{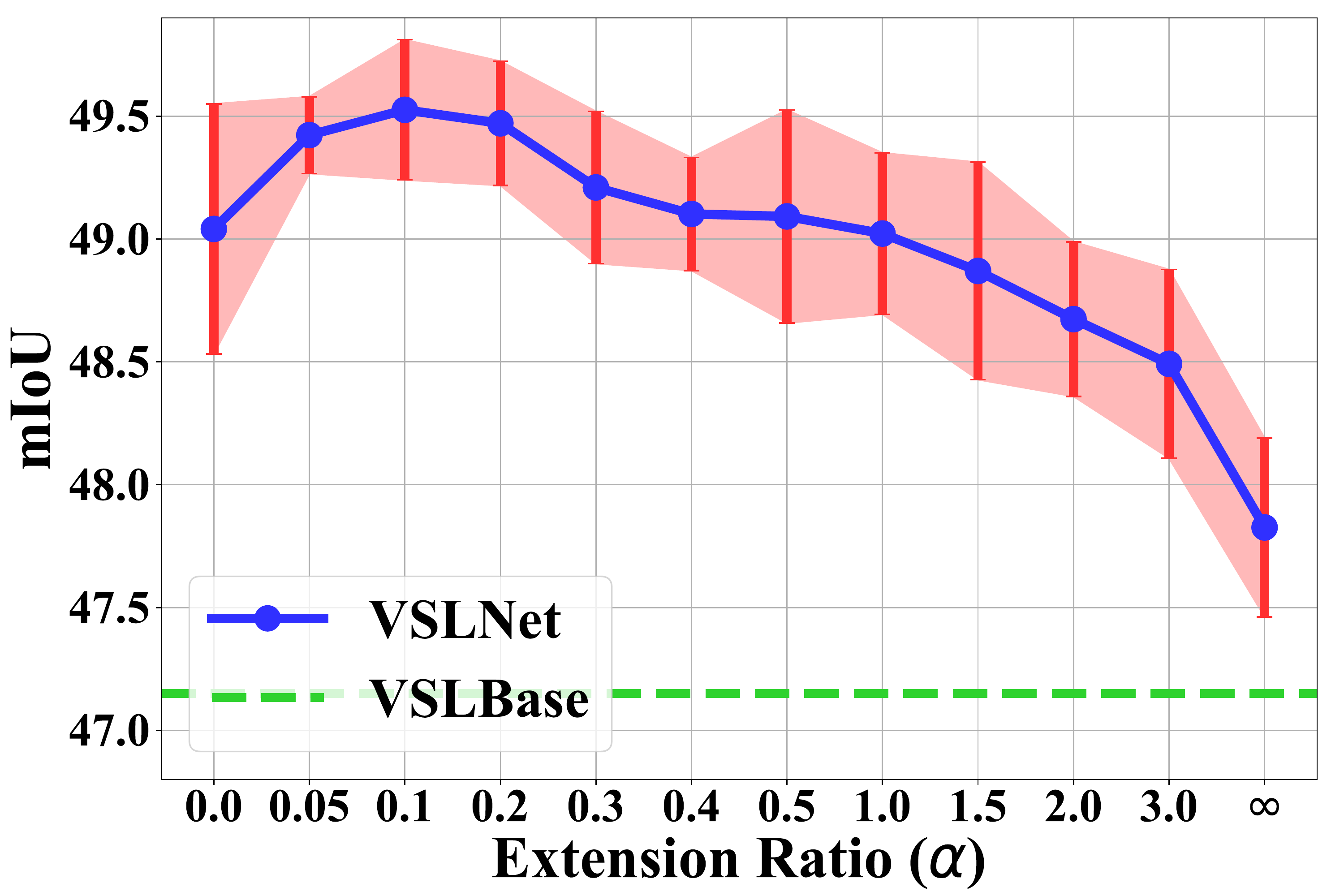}}
	\caption{\small Analysis of the impact of extension ratio $\alpha$ in Query-Guided Highlighting on Charades-STA.}
	\label{fig_charades_mask_ratio}
\end{figure}

\begin{table}[t]
    \small
    \caption{\small Results ($\%$) of VSLNet-L on TACoS using different split scales with $n=600$.}
    \vspace{-0.3cm}
	\centering
	\setlength{\tabcolsep}{3.6 pt}
	\begin{tabular}{ c  c  c  c  c  c  c }
		\toprule
        \multirow{2}{*}{Dataset} & \multirow{2}{*}{Model} & \multirow{2}{*}{Scales ($l$)} & \multicolumn{3}{c}{$\text{Rank@}1, \text{IoU}=\mu$} & \multirow{2}{*}{mIoU} \\
        & & & $\mu=0.3$ & $\mu=0.5$ & $\mu=0.7$ & \\
        \midrule
        \multirow{7}{*}{\tabincell{c}{\rotatebox{270}{TACoS$_{\text{org}}$}}} & VSLNet & - & 29.78 & 24.71 & 19.64 & 23.96 \\
        \cmidrule{2-7}
        & \multirow{6}{*}{\tabincell{c}{\rotatebox{270}{VSLNet-L}}} & 100 & 30.18 & 25.81 & 20.77 & 24.46 \\
        & & 120 & 31.13 & 26.87 & 21.19 & 25.12 \\
        & & 150 & 30.42 & 26.38 & 20.89 & 24.73 \\
        & & 200 & 30.59 & 26.07 & 21.01 & 24.61 \\
        \cmidrule{3-7}
        & & -$P_m$ & 32.04 & 27.92 & 23.28 & 26.40 \\
        & & -$U$   & 31.86 & 27.64 & 22.72 & 26.25 \\
        \midrule
        \multirow{7}{*}{\tabincell{c}{\rotatebox{270}{TACoS$_{\text{tan}}$}}} & VSLNet & - & 41.42 & 30.67 & 22.32 & 31.92 \\
        \cmidrule{2-7}
        & \multirow{6}{*}{\tabincell{c}{\rotatebox{270}{VSLNet-L}}} & 100 & 42.24 & 32.69 & 23.67 & 32.41 \\
        & & 120 & 43.39 & 33.37 & 24.19 & 33.45 \\
        & & 150 & 44.61 & 33.99 & 24.27 & 33.71 \\
        & & 200 & 43.74 & 33.67 & 23.74 & 33.50 \\
        \cmidrule{3-7}
        & & -$P_m$ & 47.11 & 36.34 & 26.42 & 36.61 \\
        & & -$U$   & 46.44 & 35.74 & 26.19 & 36.05 \\
        \bottomrule
	\end{tabular}
	\label{tab:single_multi}
\end{table}

\begin{figure}[t]
    \centering
	\subfigure[\small Charades-STA]
	{\label{fig_charades_hist}	\includegraphics[trim={0.3cm 0.2cm 0.3cm 0.3cm},clip, width=0.23\textwidth]{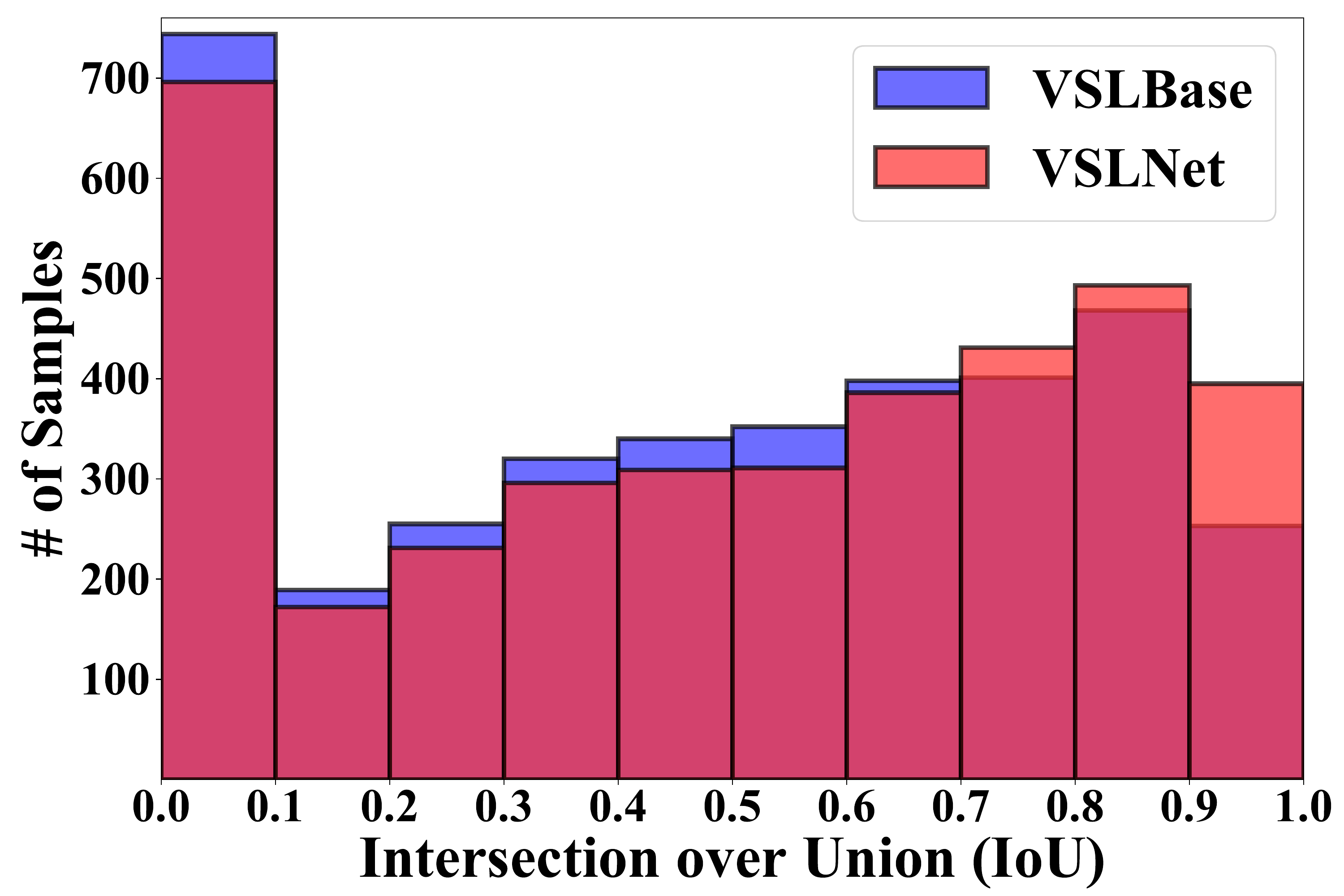}}
	\subfigure[\small ActivityNet Captions]
	{\label{fig_activitynet_hist}	\includegraphics[trim={0.3cm 0.2cm 0.3cm 0.3cm},clip, width=0.23\textwidth]{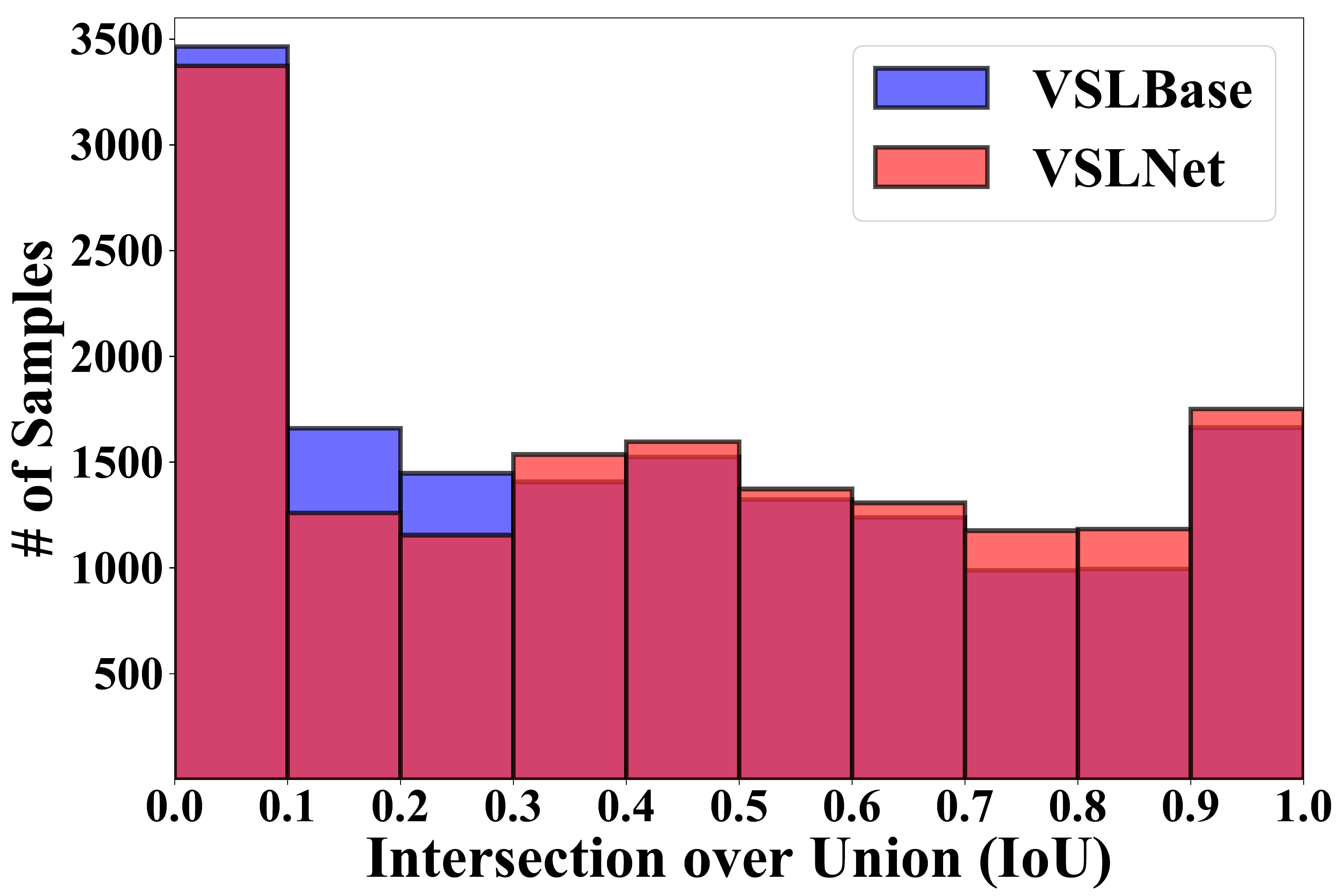}}
	\caption{\small Histograms of the number of predicted results on test set under different IoUs, on two datasets.}
	\label{fig_hist}
\end{figure}

\begin{figure*}[t]
    \centering
	\subfigure[\small Two example cases on the Charades-STA dataset]
	{\label{fig_qual_charades}	\includegraphics[width=0.95\textwidth]{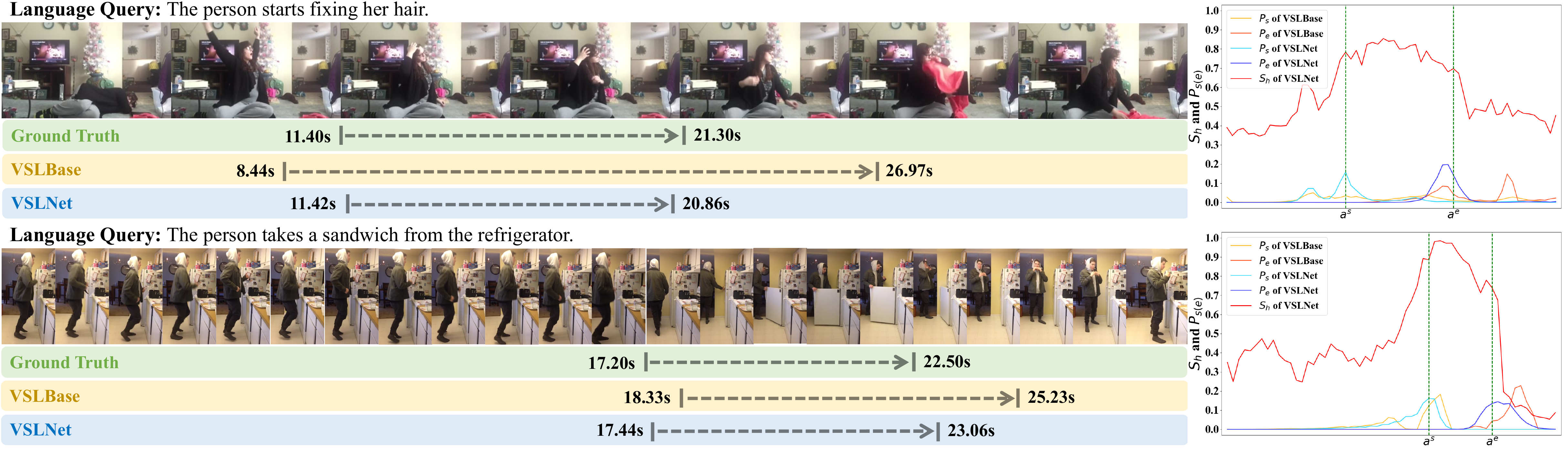}}
	\subfigure[\small Two example cases on the ActivityNet Caption dataset]
	{\label{fig_qual_activitynet}	\includegraphics[width=0.95\textwidth]{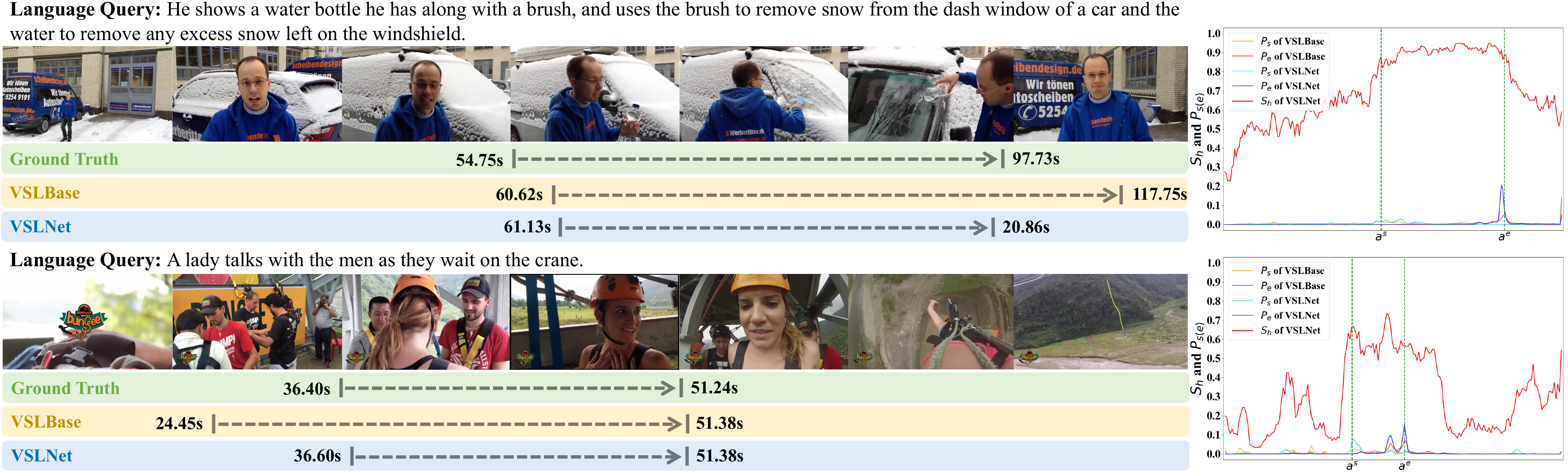}}
	\vspace{-0.3cm}
	\caption{\small Visualization of predictions by VSLBase and VSLNet. Figures on the left depict the localized results by the two models. Figures on the right show probability distributions of start/end boundaries and highlighting scores.}
	\label{fig_qual}
\end{figure*}

\begin{figure}[t]
    \centering
	\includegraphics[trim={0cm 0cm 0cm 0cm},clip,width=0.46\textwidth]{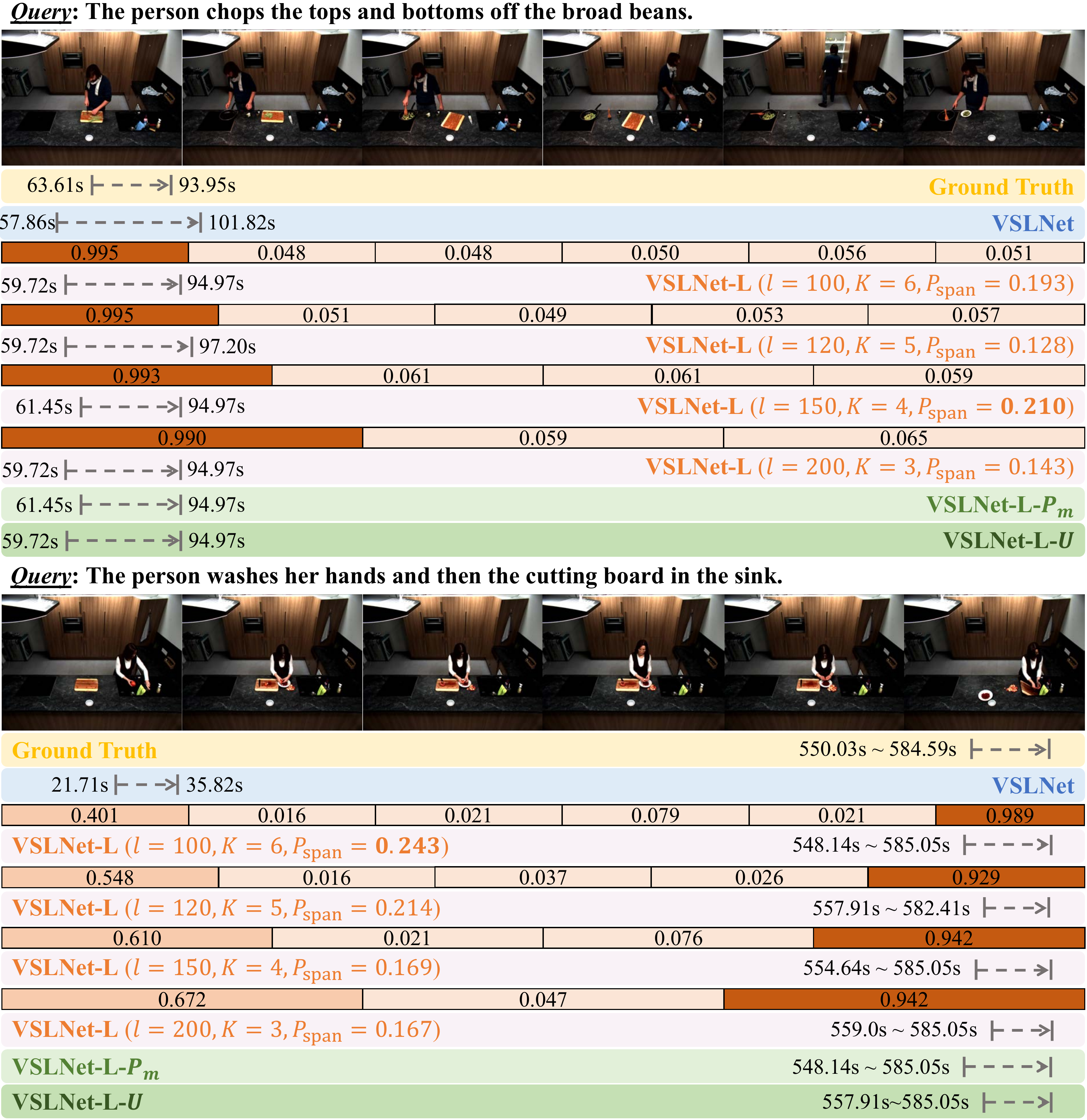}
	\vspace{-0.3cm}
	\caption{\small Visualizations of two predicted examples by VSLNet and VSLNet-L on TACoS dataset.}
	\label{fig_visualization_vslnet_l}
\end{figure}

\begin{figure}[t]
    \centering
	\subfigure[\small Charades-STA]
	{\label{fig_charades_error_time_hist}	\includegraphics[trim={0cm 0cm 0cm 0cm},clip, width=0.23\textwidth]{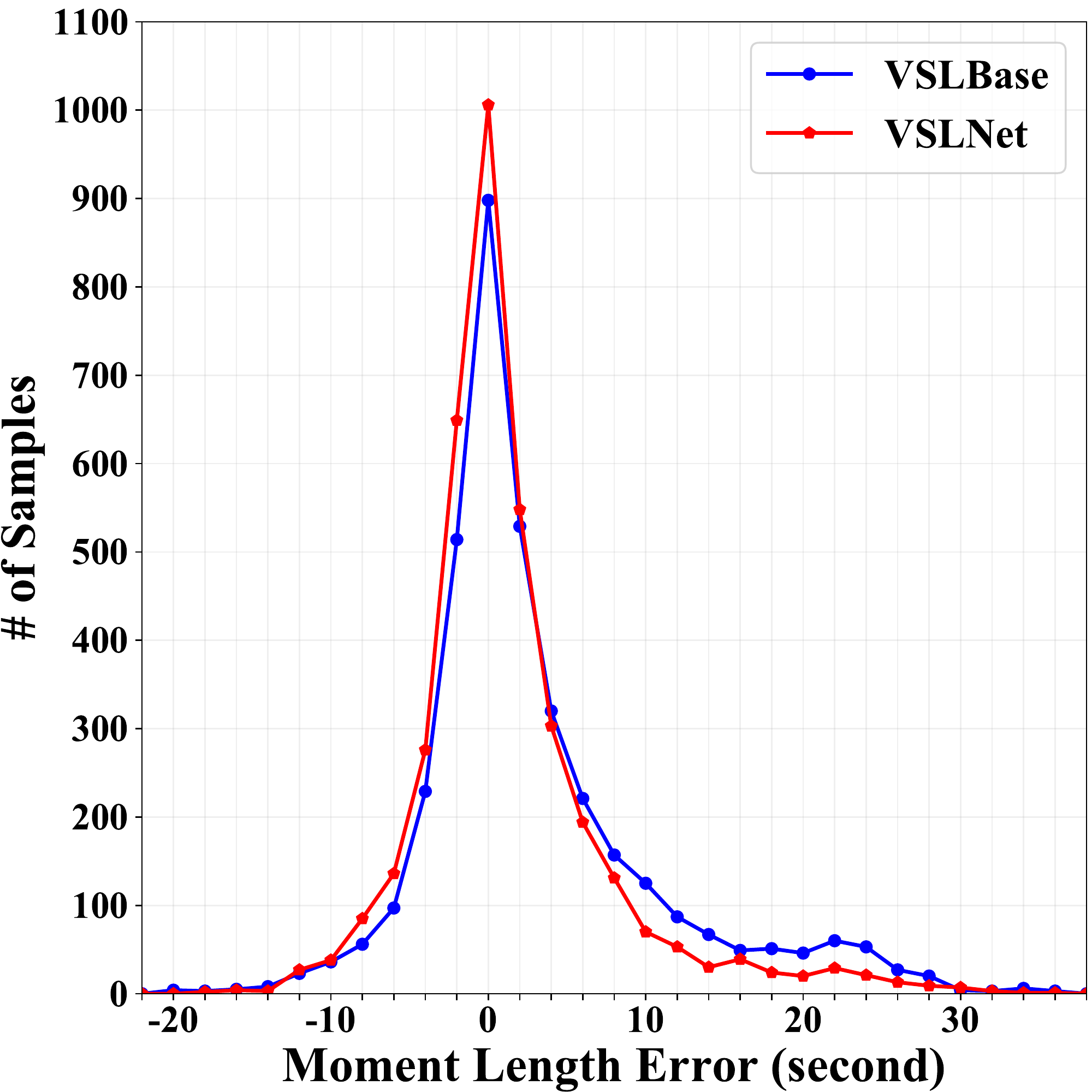}}
	\subfigure[\small ActivityNet Captions]
	{\label{fig_activitynet_error_time_hist}	\includegraphics[trim={0cm 0cm 0cm 0cm},clip, width=0.23\textwidth]{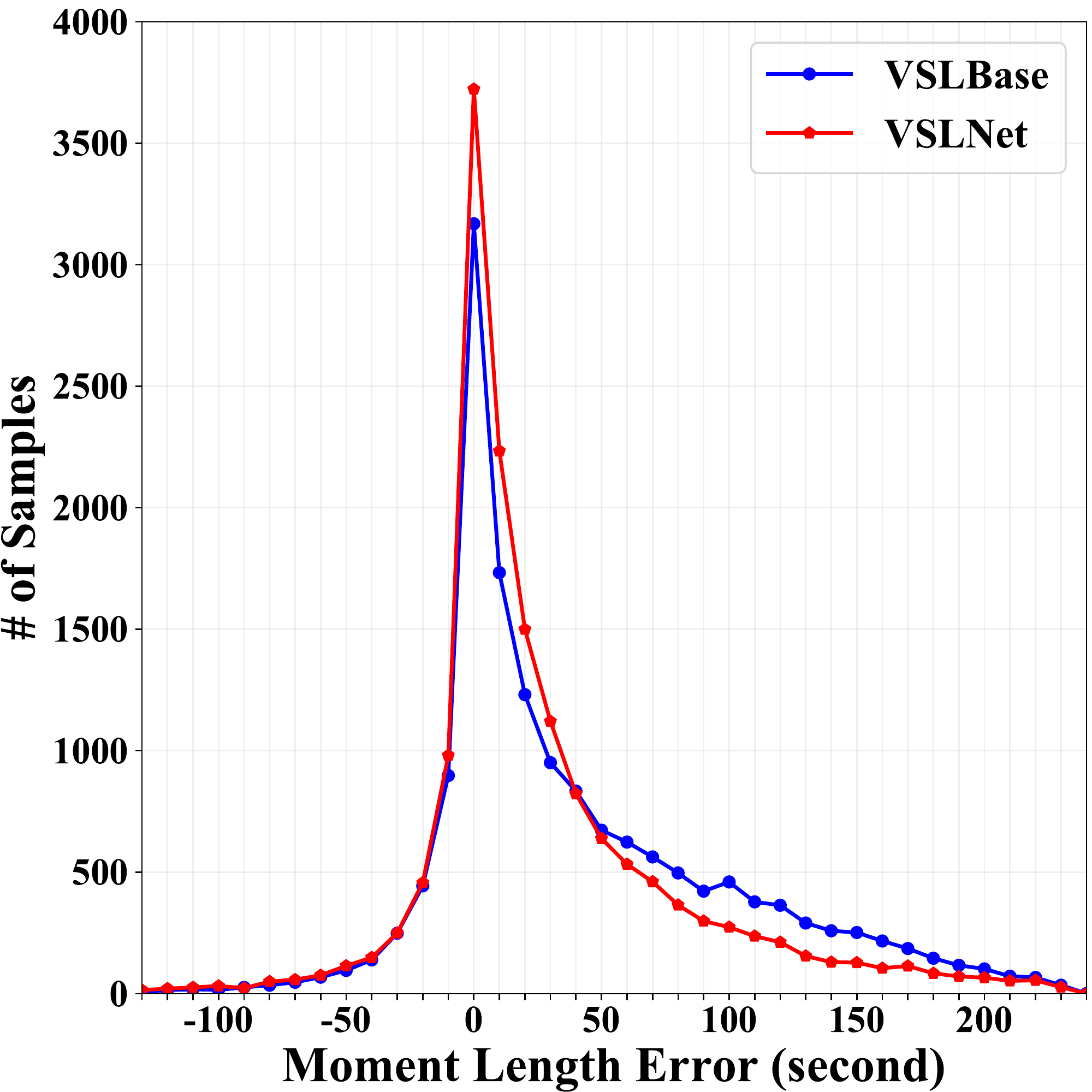}}
	\vspace{-0.3cm}
	\caption{\small Plots of moment length errors in seconds between ground truths and results predicted by VSLBase and VSLNet, respectively.}
	\label{fig_length_diff_hist}
\end{figure}

\begin{figure*}[t]
    \centering
	\subfigure[\small A failure case on the Charades-STA dataset with $\textrm{IoU}=0.11$.]
	{\label{fig_qual_charades_error}	\includegraphics[width=0.95\textwidth]{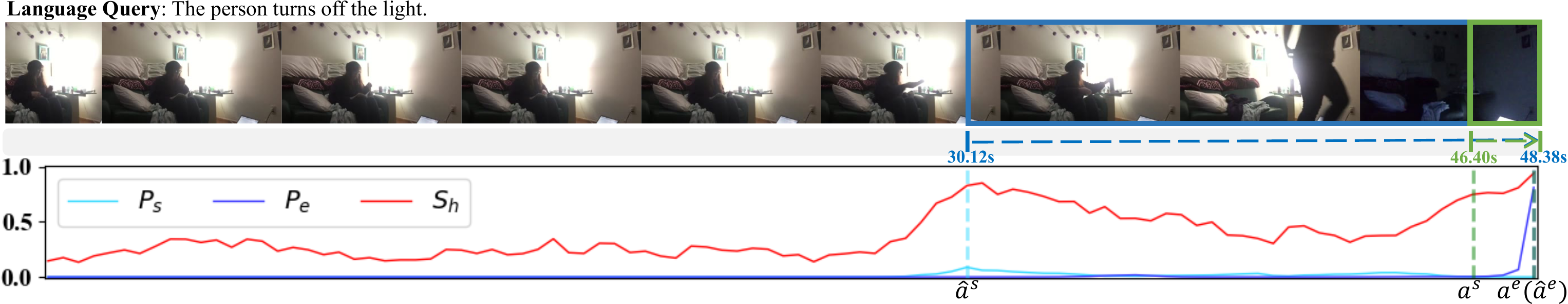}}
	\subfigure[\small A failure case on the ActivityNet Caption dataset with $\textrm{IoU}=0.17$.]
	{\label{fig_qual_activitynet_error}	\includegraphics[width=0.95\textwidth]{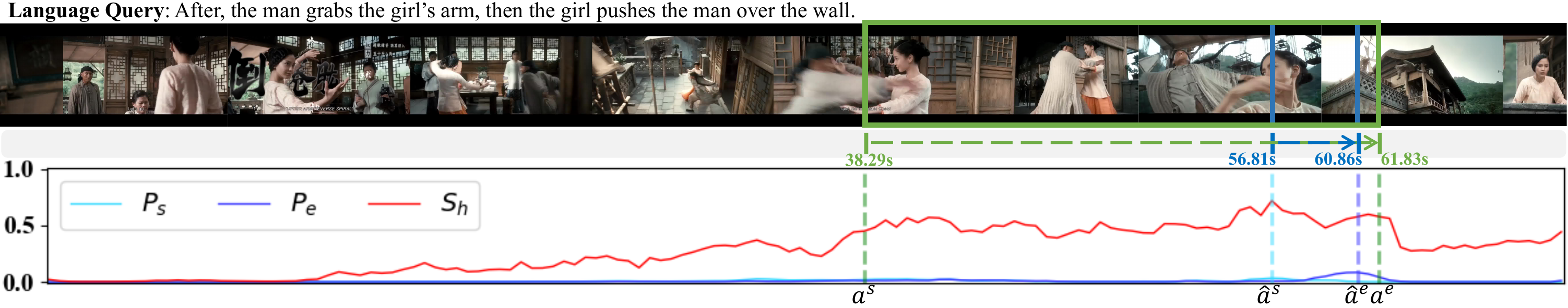}}
	\vspace{-0.3cm}
	\caption{\small Two failure examples predicted by VSLNet, $a^s/a^e$ denote the start/end boundaries of ground truth video moment, $\hat{a}^{s}/\hat{a}^{e}$ denote the start/end boundaries of predicted target moment.}
	\label{fig_qual_error}
\end{figure*}

\subsection{Performance on Videos with Different Length}\label{ssec:diff_video_len_exp}
As discussed in Section~\ref{sec:intro} and illustrated in Figure~\ref{fig_tacos_result_diff_length}, existing methods including VSLNet still underperform on NLVL with long videos. That is, the localization performance decreases dramatically along with the increase of video length. Summarized in Table~\ref{tab:stat_video_annot_length}, there are fewer videos/annotations along the increase of video length in the datasets. The relatively small number of training samples may lead to instability of the evaluated models, and performance degradation on long videos, to some extent. However, we believe that the following are the two main reasons for the performance degradation:
\begin{enumerate}
    \item Downsampling of visual features of long videos in most existing methods adversely affects localization accuracy due to information loss. As shown in Figure~\ref{fig_mean_iou_n}, sparsely downsampling video feature presentations below certain number (\eg $n<200$) would lead to dramatic performance degradation.
    \item As plotted in Figure~\ref{fig_tacos_moment_ratio}, the average normalized length of ground truth moments gradually decrease along with the increase of video length. The sparsity of moments also contributes to poor performance on long videos.
\end{enumerate}

To address this issue, VSLNet-L splits the video into multiple clip segments with different scales to simulate the multiple paragraphs in the document and maximize the chance of locating the target moment in one segment clip. 

Table~\ref{tab:duration_tacos} reports the ``mIoU'' gains of VSLNet-L on the TACoS dataset for videos with different lengths. Compared to the results of VSLNet (the best performing method without considering video length), larger improvements are observed on longer videos, which demonstrates the superiority of VSLNet-L for localizing temporal moments in long videos. For instance, VSLNet-L$^{(L)}$ achieves more than $3\%$ absolute improvements in mIoU for videos longer than $8$ minutes versus less than $2\%$ gains for videos shorter than $8$ minutes on TACoS$_{\text{org}}$. Despite the slight performance reduction on videos shorter than $2$ minutes, along with video length raises, consistent improvements are observed on TACoS$_{\text{tan}}$ for both $n=300$ $(S)$ and $600$ $(L)$, compared to VSLNet. Figure~\ref{fig_tacos_diff_length} plots the performance improvements along video lengths for better visualization. 

Results on ActivityNet Captions are reported in Table~\ref{tab:duration_activitynet}. Despite that the videos are relatively short, VSLNet-L manages to improve localization performance, with larger improvements observed on longer videos. These results show consistent superiority of VSLNet-L over VSLNet for both candidate selection strategies, on videos of different lengths.

\subsection{Ablation Studies}
In this section, we conduct ablative experiments to analyze the importance of different modules, including feature encoder and context-query attention in VSLBase. We also investigate the impact of extension ratio $\alpha$ (see Figure~\ref{fig_highlight}) in query-guided highlighting (QGH) of VSLNet, and study the impact of the multi-scale split strategy in VSLNet-L. Finally, we visually show the effectiveness of the proposed methods and discuss their limitations.
We conduct the ablation studies in this order because  VSLNet is built on top of VSLBase, and VSLNet-L is an extension of VSLNet.

\subsubsection{Module Analysis}
We first study the effectiveness of the feature encoder and context-query attention (CQA) in VSLBase by replacing them with other modules. Specifically, we use bidirectional LSTM (BiLSTM) as an alternative feature encoder. For context-query attention, we replace it by a simple method (named CAT), which concatenates each visual feature with the max-pooled query feature. 

Our feature encoder consists of Convolution + Multi-head attention + Feed-forward layers (see Section~\ref{ssec:encoder}), named as CMF here. With these alternatives, we now have four combinations, listed in Table~\ref{tab-component}. As observed from the results, CMF shows stable superiority over CAT on all metrics regardless of other modules; CQA surpasses CAT whichever feature encoder is used. This study indicates that CMF and CQA are more effective.

Table~\ref{tab-component-delta} reports the performance gains of different modules over ``$\textrm{Rank@}1, \textrm{IoU}=0.7$'' metric. The results show that replacing CAT with CQA leads to larger improvements, compared to replacing BiLSTM by CMF. This observation suggests that CQA plays a more important role in our model. Specifically, keeping CQA, the absolute gain is $1.61\%$ by replacing the encoder module. Keeping CMF, the gain of replacing the attention module is $3.09\%$.

Figure~\ref{fig_attn_score} displays the matrix of similarity score between visual and language features in the context-query attention (CQA) module ($\mathcal{S}\in\mathbb{R}^{n\times m}$ in Section~\ref{ssec:cqa}). This figure shows visual features are more relevant to the verbs and their objects in the query sentence. For example, the similarity scores between visual features and ``\textit{eating}'' (or ``\textit{sandwich}'') are higher than that of other words. We believe that verbs and their objects are more likely to be used to describe video activities. Our observation is consistent with Ge \etal~\cite{ge2019mac}, where \textit{verb-object} pairs are extracted as semantic activity concepts. In contrast, these concepts are automatically captured by the CQA module in our method.

\subsubsection{The Impact of Extension Ratio in QGH}
VSLNet introduces a query-guided highlighting (QGH) module on top of VSLBase to address the technical gaps between video and text. The QGH guides the VSLNet to search for the target moment within a longer highlighted region controlled by the extension ratio $\alpha$. We now investigate the impact of the extension ratio $\alpha$ in the QGH module on the Charades-STA dataset. We evaluated $12$ different values of $\alpha$ from $0.0$ to $\infty$ in experiments. $0.0$ represents no answer span extension, and $\infty$ means that the entire video is regarded as foreground. The results for various $\alpha$'s are plotted in Figure~\ref{fig_charades_mask_ratio}. It shows that query-guided highlighting consistently contributes to performance improvements, regardless of $\alpha$ values, \ie from $0$ to $\infty$. Along with $\alpha$ raises, the performance of VSLNet first increases and then gradually decreases. The optimal performance appears between $\alpha=0.05$ and $0.2$ over all metrics. 

Note that, when $\alpha=\infty$, which is equivalent to no region is highlighted as a coarse region to locate target moment, VSLNet remains better than VSLBase. Shown in Figure~\ref{fig_qgh}, when $\alpha=\infty$, QGH effectively becomes a straightforward concatenation of sentence representation with each of the visual features. The resultant feature remains helpful for capturing semantic correlations between vision and language. In this sense, this function can be regarded as an approximation or simulation of the traditional multimodal matching strategy~\cite{Hendricks2017LocalizingMI,Gao2017TALLTA,Liu2018AMR}.

\subsubsection{Single \textit{vs} Multi-scale Split-and-Concat}\label{sssec:single_multi_scale}
VSLNet-L further introduces a multi-scale split-and-concat strategy to address performance degradation on long videos. Here, we study the impact of the multi-scale split on the TACoS dataset with $n=600$, against the single-scale split. We evaluate $4$ different values of single-scale, \ie $l\in\{100, 120, 150, 200\}$. The multi-scale mechanism is jointly trained with the four scales. The results are summarized in Table~\ref{tab:single_multi}. Compared to VSLNet, split-and-concat strategy in VSLNet-L consistently contributes to performance improvements, regardless of the $l$ value. The best single-scale $l$ is $120$ for TACoS$_{\text{org}}$, and $150$ for TACoS$_{\text{tan}}$. Compared to VSLNet-L with single-scale, VSLNet-L-$P_m$ (-$U$) further improves all metrics significantly. The multi-scale split-and-concat mechanism not only alleviates the issue of target moment truncation but also captures contextual information in the video; both improve the generalization ability of VSLNet-L.

\subsubsection{Qualitative Analysis}\label{sssec:qualitative}
Figure~\ref{fig_hist} shows the histograms of predicted results of VSLBase and VSLNet on test sets of the Charades-STA and ActivityNet Captions datasets. Results indicate that VSLNet beats VSLBase by having more samples in the high IoU ranges, \eg $\textrm{IoU}\geq 0.7$ on the Charades-STA dataset. More predicted results of VSLNet are distributed in the high IoU ranges for the ActivityNet Caption dataset. This result demonstrates the effectiveness of the query-guided highlighting (QGH) strategy.

We show two examples of VSLBase and VSLNet in Figures~\ref{fig_qual_charades} and~\ref{fig_qual_activitynet} from the Charades-STA and ActivityNet Captions datasets, respectively. From the two figures, the localized moments by VSLNet are closer to the ground truth than that by VSLBase. Meanwhile, the start and end boundaries predicted by VSLNet are roughly constrained in the highlighted regions $S_{\textrm{h}}$, computed by QGH. 

Figure~\ref{fig_visualization_vslnet_l} depicts two predicted examples from the TACoS dataset as case studies. The localized moments by VSLNet-L are more accurate and closer to the ground truth moment than that of VSLNet. Both figures show the results of VSLNet-L are constrained in the positive clip segments, \ie the clip segment that contains ground truth. In the second example, VSLNet does not capture the concept ``the cutting board in the sink'' and focuses on retrieving ``washes'' action only, which leads to an error prediction. In contrast, VSLNet-L correctly understands both ``washes'' and the position of ``cutting board'', leading to the correct prediction.

\subsubsection{Error Analysis}\label{sssec:error}
We further study the error patterns of predicted moment lengths on VSLBase and VSLNet, as shown in Figure~\ref{fig_length_diff_hist}. The differences between moment lengths of ground truths and predicted results are measured. A positive length difference means the predicted moment is longer than the corresponding ground truth, while a negative means shorter. Figure~\ref{fig_length_diff_hist} shows that VSLBase tends to predict longer moments, \eg more samples with length error larger than $4$ seconds in Charades-STA or 30 seconds in ActivityNet. On the contrary, constrained by QGH, VSLNet tends to predict shorter moments, \eg more samples with length error smaller than $-4$ seconds in Charades-STA or $-20$ seconds in ActivityNet Caption. This observation is helpful for future research on adopting the span-based QA framework for NLVL.

In addition, we also exam failure cases (with IoU predicted by VSLNet lower than $0.2$) shown in Figure~\ref{fig_qual_error}. In the first case, as illustrated by Figure~\ref{fig_qual_charades_error}, we observe an action that a person turns towards to the lamp and places an item there. The QGH falsely predicts the action as the beginning of the moment ``turns off the light''. The second failure case involves multiple actions in a query, as shown in Figure~\ref{fig_qual_activitynet_error}. QGH successfully highlights the correct region by capturing the temporal information of two different action descriptions in the given query. However, it assigns ``pushes'' with a higher confidence score than ``grabs''. Thus, VSLNet only captures the region corresponding to the ``pushes'' action, due to its confidence score.

\section{Conclusion}\label{sec:conclusion}
In this paper, we revisit the NLVL task and propose to solve it with a multimodal span-based QA framework by considering a video as a text passage. We show that adopting a standard span-based QA framework, VSLBase, can achieve promising results on the NLVL task. However, there are two major differences between video and text in the standard span-based QA framework, limiting the performance of VSLBase. To address the differences, we then propose VSLNet, which introduces a simple and effective strategy named query-guided highlighting (QGH), on top of VSLBase. With QGH, VSLNet is guided to search for answers within a predicted coarse region. The effectiveness of VSLNet (and VSLBase) is demonstrated with experiments on three datasets. The results indicate that it is promising to explore the span-based QA framework to address NLVL problems. Moreover, we have observed that the existing methods including VSLNet suffer from the performance degradation issue on long videos. To address this issue, we further extend VSLNet by regarding a long video as a document with multiple paragraphs. We adopt the concept of MPQA and propose a multi-scale split-and-concat network to capture contextual information in a long video by partitioning the long video multiple times with different clip lengths. Extensive experiments demonstrate that VSLNet-L prevents performance degradation on long videos effectively and advances the state-of-the-art for NLVL on three benchmark datasets.

\bibliographystyle{IEEEtran}
\bibliography{IEEEtran}

\end{document}

%% file: math_commands.tex
\usepackage{amsmath,amsfonts,bm}

\def\eqref#1{equation~\ref{#1}}

\def\1{\bm{1}}

\DeclareMathAlphabet{\mathsfit}{\encodingdefault}{\sfdefault}{m}{sl}
\SetMathAlphabet{\mathsfit}{bold}{\encodingdefault}{\sfdefault}{bx}{n}

